\documentclass[journal]{IEEEtran}

\usepackage[pdftex]{graphicx}
\usepackage{amsmath,color,bm,amssymb,mathrsfs,subfigure,booktabs,cite,url,mathrsfs}

\begin{document}

\setlength{\abovedisplayskip}{4pt}
\setlength{\belowdisplayskip}{4pt}
\setlength{\arraycolsep}{0.1em}
\setlength{\aboverulesep}{0.3pt}
\setlength{\belowrulesep}{0.7pt}

\title{
Rhythm Transcription of Polyphonic Piano Music\\ Based on Merged-Output HMM for Multiple Voices
}

\author{
Eita Nakamura,~\IEEEmembership{Member,~IEEE,}
Kazuyoshi Yoshii,~\IEEEmembership{Member,~IEEE,}
and~Shigeki Sagayama,~\IEEEmembership{Member,~IEEE}
\thanks{E.~Nakamura and K.~Yoshii are with the Graduate School of Informatics, Kyoto University, Kyoto 606-8501, Japan (e-mail: {\tt enakamura@sap.ist.i.kyoto-u.ac.jp}, {\tt yoshii@kuis.kyoto-u.ac.jp}). E.~Nakamura is supported by the JSPS research fellowship (PD).}
\thanks{S.~Sagayama is with the Graduate School of Advanced Mathematical Sciences, Meiji University, Nakano, Tokyo 164-8525, Japan (e-mail: {\tt sagayama@meiji.ac.jp}).}
%\thanks{Manuscript received XX, YY; revised XX, YY.}
}

\markboth{IEEE/ACM TRANSACTIONS ON AUDIO, SPEECH, AND LANGUAGE PROCESSING, VOL. XX, No. YY, 2017}
{Nakamura \MakeLowercase{\textit{et al.}}: Rhythm Transcription of Polyphonic MIDI Performances}



\maketitle

\begin{abstract}
In a recent conference paper, we have reported a rhythm transcription method based on a merged-output hidden Markov model (HMM) that explicitly describes the multiple-voice structure of polyphonic music. This model solves a major problem of conventional methods that could not properly describe the nature of multiple voices as in polyrhythmic scores or in the phenomenon of loose synchrony between voices. In this paper we present a complete description of the proposed model and develop an inference technique, which is valid for any merged-output HMMs for which output probabilities depend on past events. We also examine the influence of the architecture and parameters of the method in terms of accuracies of rhythm transcription and voice separation and perform comparative evaluations with six other algorithms. Using MIDI recordings of classical piano pieces, we found that the proposed model outperformed other methods by more than 12 points in the accuracy for polyrhythmic performances and performed almost as good as the best one for non-polyrhythmic performances. This reveals the state-of-the-art methods of rhythm transcription for the first time in the literature. Publicly available source codes are also provided for future comparisons.
\end{abstract}


% \fi
\IEEEpeerreviewmaketitle

%%%%%%%%%%%%%%%%%%%%%%%%%%%%%%%%%%%%%%%%%%%%%%
\section{Introduction}
%%%%%%%%%%%%%%%%%%%%%%%%%%%%%%%%%%%%%%%%%%%%%%

\IEEEPARstart{M}{usic} transcription is one of the most challenging problems in music information processing.
To obtain music scores, we need to extract pitch information from music audio signals.
Recently pitch analysis for polyphonic (e.g.\ piano) music has been receiving much attention \cite{Klapuri2006,Benetos2013}.
To solve the other part of the transcription problem, many studies have been devoted to so-called rhythm transcription, that is, the problem of recognising quantised note lengths (or note values) of the musical notes in MIDI performances \cite{LonguetHiggins1987,Pressing1993,Chowning1984,Temperley1999,Desain1989,Otsuki2002,Takeda2002,Takeda2007,Raphael2001,Hamanaka2003,Cemgil2003,Temperley2009,Tanji2008,Tsuchiya2013,Takamune2014,Nakamura2016,NakamuraSMC2016}.

Since early studies in the 1980s, various methods have been proposed for rhythm transcription.
As we explain in detail in Sec.~\ref{sec:RelatedWork}, the general trend has shifted to using machine learning techniques to capture what natural music scores are and how music performances fluctuate in time.
One of the models most frequently used in recent studies \cite{Otsuki2002,Takeda2002,Raphael2001,Hamanaka2003,Cemgil2003,Takeda2007,Nakamura2016,NakamuraSMC2016} is the hidden Markov model (HMM) \cite{Rabiner1989}.
In spite of its importance and about 30 years of history, however, little comparative evaluations on rhythm transcription have been reported in the literature and the state-of-the-art method has not been known.

Rhythm transcription also raises challenging problems of representing and modelling scores and performances for polyphonic music.
This is because a polyphonic score has multilayer structure, where concurrently sounding notes are grouped into several streams, or {\it voices}\footnote{In this paper, a `voice' means a unit stream of musical notes that can contain chords. The score in Fig.~\ref{fig:PreviousModels}, for example, has two voices corresponding to the left and right hands.}.
As explained in Sec.~\ref{sec:RelatedWork}, a conventional way of representing a polyphonic score as a linear sequence of chords \cite{Takeda2007} may not retain sequential regularities within voices, such as those in polyrhythmic scores, nor it can capture the loose synchrony between voices \cite{Heijink2000,Gingras2011} in polyphonic performances.
Therefore solutions to explicitly describe the multiple-voice structure must be sought.

\begin{figure}[t]
\begin{center}
{\includegraphics[clip,width=1.\columnwidth]{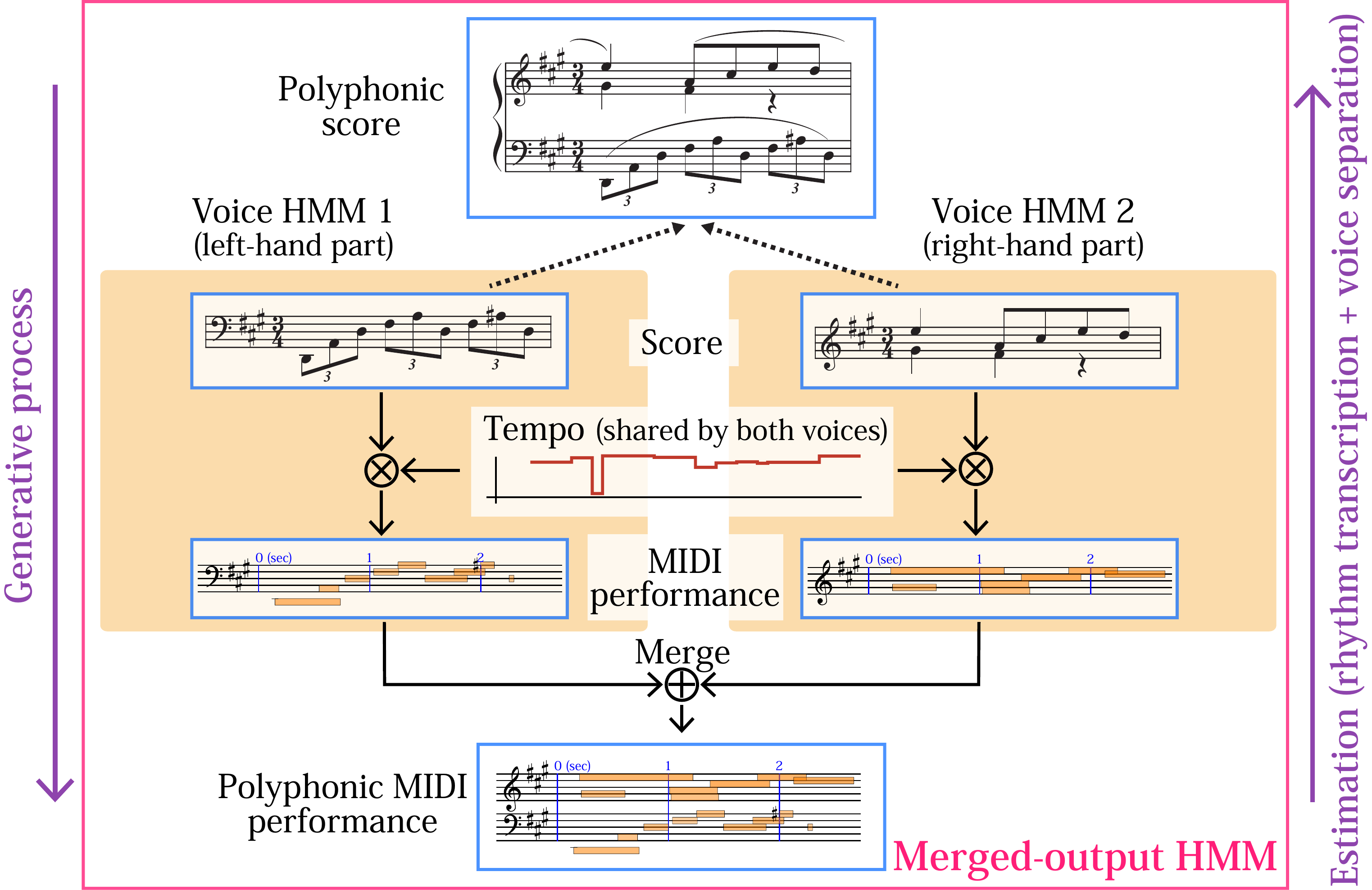}}
\end{center}
\vspace{-5mm}
\caption{Overview of the proposed model describing the generation of polyphonic performances.}
\label{fig:Overview}
\vspace{-3mm}
\end{figure}
From this point of view, in a recent conference \cite{NakamuraSMC2016}, we reported a statistical model that can describe the multiple-voice structure of polyphonic music.
The model is based on the merged-output HMM \cite{ICMC,ISMIR2014}, which describes polyphonic performances as merged outputs from multiple component HMMs, called {\it voice HMMs}, each of which describes the generative process of music scores and performances of one voice (Fig.~\ref{fig:Overview}).
It was confirmed that the model outperformed conventional HMM-based methods for transcribing polyrhythmic performances.

The purpose of this paper is to discuss in detail the merged-output HMM and its inference technique.
Due to the large size of the state space and the complex dependencies between variables, the standard Viterbi algorithm or its refined version \cite{ICMC} cannot be applied and a new inference technique is necessary.
This problem typically arises when a voice HMM is an autoregressive HMM, which is commonly used as music score/performance models where output probabilities of events (e.g.~pitch, time, etc.) depend on past events.
Using a trick of introducing an auxiliary variable to trace the history of output symbols similarly as in Ref.~\cite{ISMIR2014}, we develop an inference technique that can work in a practical computer environment and could be applied for any merged-output HMMs with autoregressive voice HMMs.

We provide a complete description of the proposed model and examine the influence of its architecture and parameters.
First, we explain details omitted in the previous paper including the description of the chord model and a switch of a coupling parameter between voice HMMs depending on pitch contexts.
The effects are examined in terms of accuracies.
Second, the determination of model parameters based on supervised learning is discussed and the influence of parameters of the performance model is investigated.
Finally, a feature of the proposed method is its simultaneous voice separation and rhythm recognition.
We examine this effect by evaluating accuracies of both voice separation and rhythm recognition and comparing with a cascading algorithm that performs voice separation first and then recognises rhythm.

Another contribution of this paper is to present results of systematic comparative evaluations to find the state-of-the-art method.
In addition to two HMM-based methods \cite{Otsuki2002,Takeda2002,Raphael2001,Hamanaka2003} previously tested in Ref.~\cite{NakamuraSMC2016}, we tested frequently cited methods and theoretically important methods whose source codes were available: Connectionist Quantizer \cite{Desain1989}, Melisma Analyzers (version 1 \cite{Temperley1999} and version 2 \cite{Temperley2009}) and two-dimensional (2D) probabilistic context-free grammar (PCFG) model \cite{Kameoka2012,Tsuchiya2013,Takamune2014}.
An evaluation measure for rhythm transcription, which is briefly sketched in Ref.~\cite{NakamuraSMC2016}, is explained in full detail together with its calculation algorithm.

We make public the source codes for the best models found (the proposed model and other two HMMs) as well as the evaluation tool to enable future comparisons \cite{DemoPage}.
We hope that these materials would encourage researchers interested in music transcription and symbolic music processing.

%%%%%%%%%%%%%%%%%%%%%%%%%%%%%%%%%%%%%%%%%%%%%%
\section{Related Work}\label{sec:RelatedWork}
%%%%%%%%%%%%%%%%%%%%%%%%%%%%%%%%%%%%%%%%%%%%%%

Previous studies on rhythm transcription are reviewed in this section.
The purpose is two-fold: First, we describe the historical development of models for rhythm transcription, some of which form bases of our model and some are subjects of our comparative evaluation.
Second, we review how polyphony has been treated in previous studies in rhythm transcription and related fields and explain in details the motivations for explicitly modelling multiple voices.
Part of discussions in Secs.~\ref{sec:StatisticalMethods} and \ref{sec:PolyphonicExtension} and the figures are quoted from Ref.~\cite{NakamuraSMC2016} to make this section more informative and self-contained.

\subsection{Early Studies}

Until the late 1990s, studies on rhythm transcription used models describing the process of quantising note durations and/or recognising the metre structure.
Longuet-Higgins \cite{LonguetHiggins1987} developed a method for estimating the note values and the metre structure simultaneously by recursively subdividing a time interval into two or three almost equally spaced parts that are likely to begin at note onsets.
A similar method of dividing a time interval using template grids and an error function of onsets and inter-onset intervals (IOIs) has also been proposed \cite{Pressing1993}.
Methods using preference rules for the ratios of quantised note durations have been developed by Chowning et al.\ \cite{Chowning1984} and Temperley et al.\ \cite{Temperley1999}.
Desain et al.\ \cite{Desain1989} proposed a connectionist approach that iteratively converts note durations so that adjacent durations tend to have simple integral ratios.

Despite some successful results, these methods have limitations in principle.
First, they use little or no information about sequential regularities of note values in music scores.
Since there are many logically possible sequences of note values, such sequential regularities are important clues to finding the one that is most likely to appear in actual music scores.
Second, tendencies of temporal fluctuations in human performances are described only roughly.
In particular, the chord clustering---that is, the identification of notes whose onsets are exactly simultaneous in the score---is handled with thresholding or is not treated at all.
Finally, the parameters of most of those algorithms are tuned manually and optimisation methods have not been developed.
This means that one cannot utilise a data set of music scores and performances to learn the parameters, or only inefficient optimisation methods like grid search can be applied.

\subsection{Statistical Methods}\label{sec:StatisticalMethods}

Since around the year 2000, it has become popular to use statistical models, which enable us to utilise the statistical nature of music scores and performances.
Usually two models, one describing the probability of a score ({\it score model}) and the other describing the probability of a performance given a score ({\it performance model}), are combined as a Bayesian model, and rhythm transcription can be formulated as maximum a posteriori estimation.
Below we review representative models for rhythm transcription.
We here consider only monophonic performances; polyphonic extensions are described in Sec.~\ref{sec:PolyphonicExtension}.

\begin{figure}[t]
\begin{center}
{\includegraphics[clip,width=0.9\columnwidth]{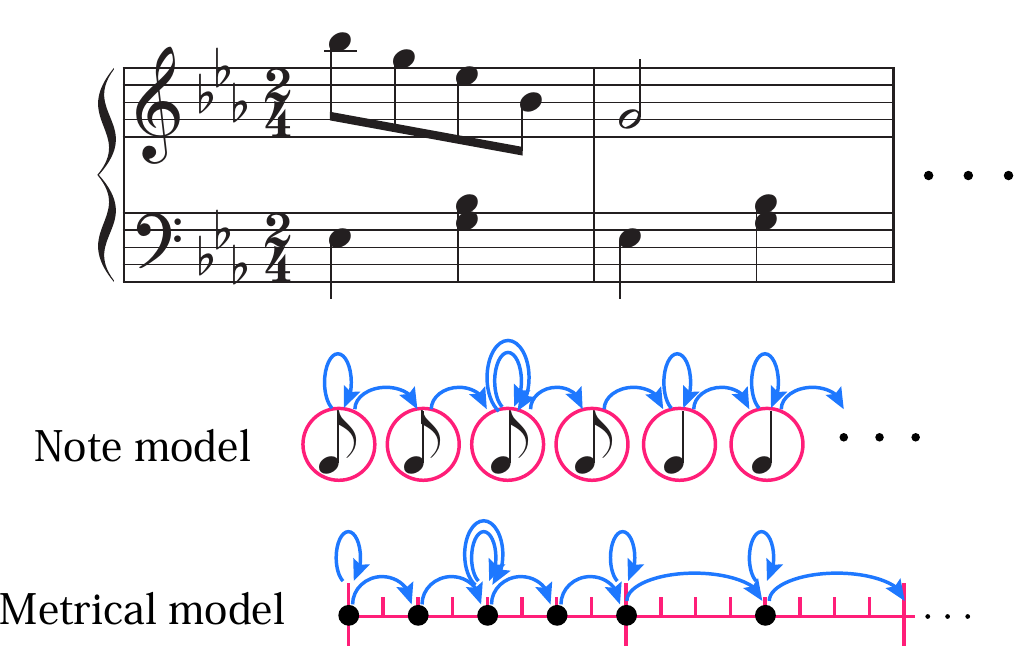}}
\end{center}
%\vspace{-5mm}
\caption{Two different representations of a music score in previously proposed HMMs.}
\label{fig:PreviousModels}
%\vspace{-3mm}
\end{figure}
In one class of HMMs for rhythm transcription, which we call {\it note HMMs}, a score is represented as a sequence of note values and described with a Markov model (Fig.~\ref{fig:PreviousModels}) \cite{Otsuki2002,Takeda2002}.
To describe the temporal fluctuations in performances, one introduces a latent variable corresponding to a (local) tempo that is also described with a Markov model.
An observed duration is described as a product of the note value and the tempo that is exposed to noise of onset times.

In another class of HMMs, which we call {\it metrical HMMs}, a different description is used for the score model \cite{Raphael2001,Hamanaka2003,Cemgil2003}.
Instead of a Markov model of note values, a Markov process on a grid space representing beat positions of a unit interval, such as a bar, is considered (Fig.~\ref{fig:PreviousModels}).
The note values are given as differences between successive beat positions.
Incorporation of the metre structure is an advantage of metrical HMMs.

{\it PCFG models} have also been proposed \cite{Tanji2008,Tsuchiya2013}.
As in \cite{LonguetHiggins1987}, a time interval in a score is recursively divided into shorter intervals until those corresponding to note values are obtained, and probabilities describe what particular divisions are likely.
As an advantage, modifications of rhythms by inserting (splitting) notes can be naturally described with these models.

\subsection{Polyphonic Extensions}\label{sec:PolyphonicExtension}

The note HMM has been extended to handle polyphonic performances \cite{Takeda2007}.
This is done by representing a polyphonic score as a linear sequence of chords or, more precisely, {\it note clusters} consisting of one or more notes.
Such score representation is also familiarly used for music analysis \cite{Conklin2002} and score-performance matching \cite{Cont,Nakamura2015}.
Chordal notes can be represented as self-transitions in the score model (Fig.~\ref{fig:PreviousModels}) and their IOIs can be described with a probability distribution with a peak at zero.
Polyphonic extension of metrical HMMs is possible in the same way.

\begin{figure}[t]
\begin{center}
\includegraphics[clip,width=1.\columnwidth]{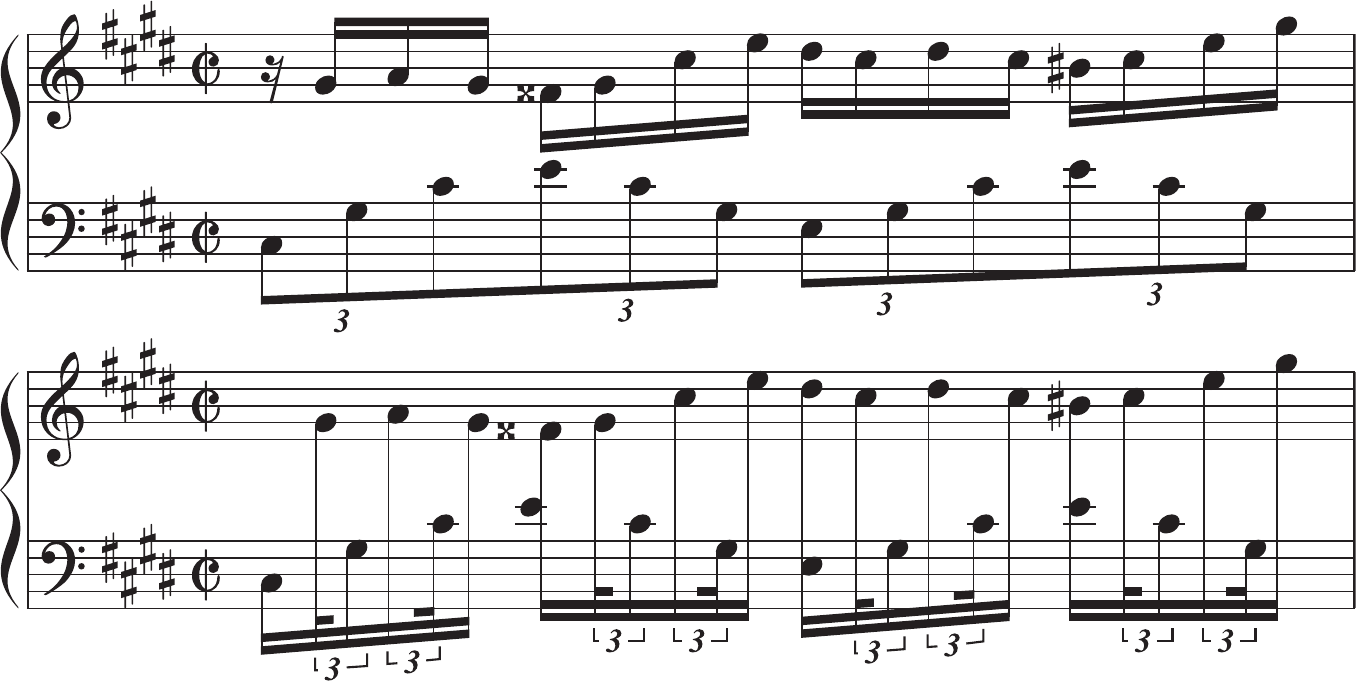}
\end{center}
%\vspace{-5mm}
\caption{A 3 against 4 polyrhythmic passage (top; Chopin's Fantaisie Impromptu) represented as a sequence of note clusters (bottom).}
\label{fig:HomophonizationExample}
%\vspace{-1mm}
\end{figure}
Although this simplified representation of polyphonic scores is logically possible, there are instances in which score and performance models based on this representation cannot describe the nature of polyphonic music appropriately.
First, complex polyphonic scores such as polyrhythmic scores are forced to have unrealistically small probabilities.
This is because such scores consist of rare rhythms in the simplified representation even if the component voices have common rhythms (Fig.~\ref{fig:HomophonizationExample}).
Second, the phenomenon of loose synchrony between voices (e.g.\ two hands in piano performances \cite{Heijink2000}), called {\it voice asynchrony}, cannot be described.
For example, the importance of incorporating the multiple-voice structure in the presence of voice asynchrony is well investigated in studies on score-performance matching \cite{Heijink2000,Gingras2011}.

To describe the multiple-voice structure of polyphonic scores, an extension of the PCFG model called 2D PCFG model has been proposed \cite{Kameoka2012,Tsuchiya2013}.
This model describes, in addition to the divisions of a time interval, duplications of intervals into two voices.
Unfortunately, a tractable inference algorithm could not be obtained for the model, and the correct voice information had to be provided for evaluations.
In a recent report, Takamune et al.\ \cite{Takamune2014} state that this problem is solved using a generalised LR parser.
However, as we shall see in Sec.~\ref{sec:Comparison}, their algorithm often fails to output results and the computational cost is quite high.

\subsection{Merged-Output HMM}\label{sec:MOHMM}

\begin{figure}[t]
\begin{center}
{\includegraphics[clip,width=1.\columnwidth]{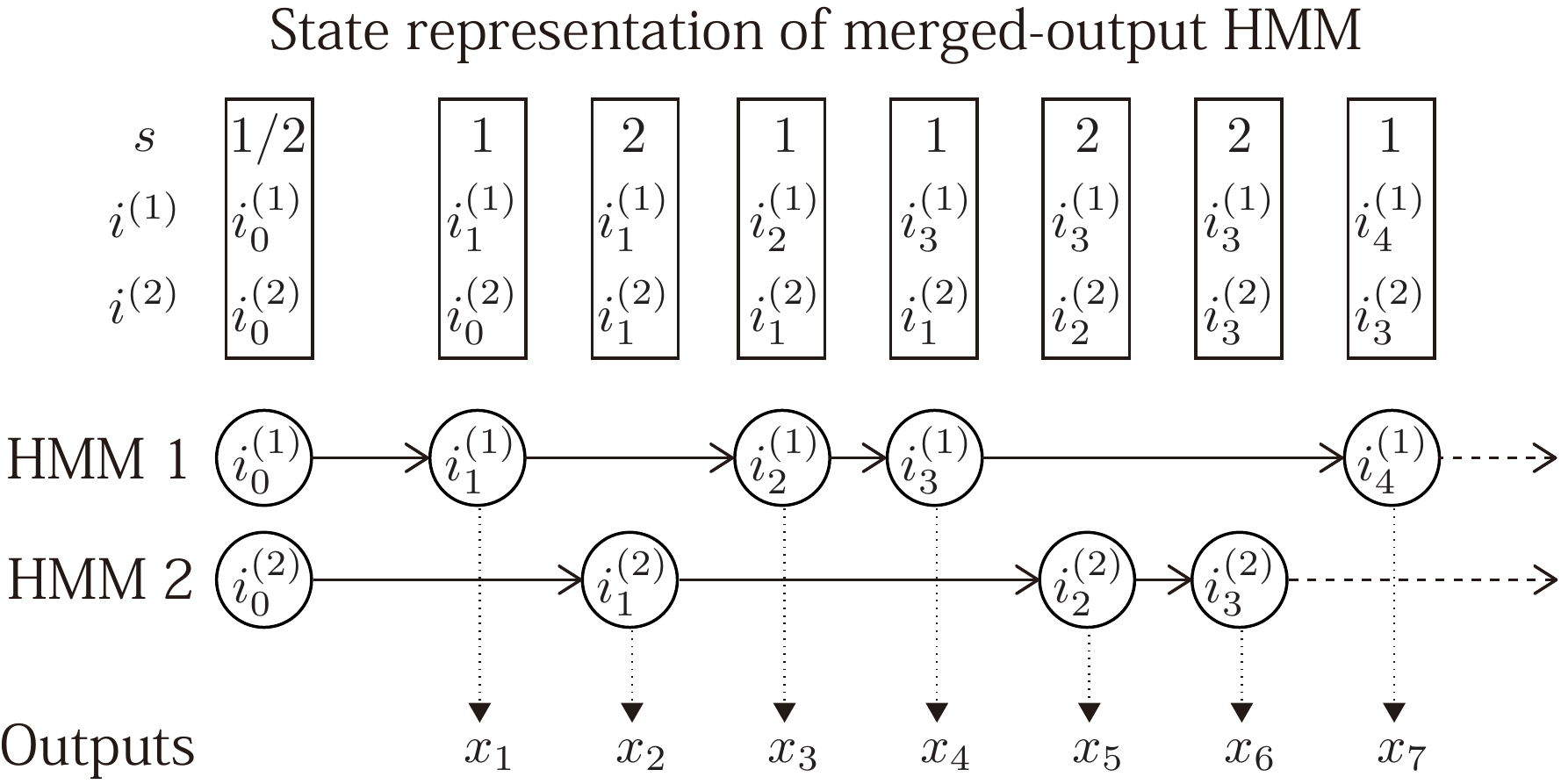}}
\end{center}
%\vspace{-5mm}
\caption{A schematic illustration of the merged-output HMM. The symbols $i^{(1)}_0$ and $i^{(2)}_0$ represent auxiliary states to define the initial transitions.}
\label{fig:MergedOutputHMM}
%\vspace{-3mm}
\end{figure}
Based on the fact that HMM is effective for monophonic music \cite{Otsuki2002,Takeda2002,Takeda2007,Raphael2001,Hamanaka2003,Cemgil2003}, an HMM-based model that can describe multiple-voice structure of symbolic music, called merged-output HMM, has been proposed \cite{ICMC,ISMIR2014}.
In the model, each voice is described with an HMM, called a voice HMM, and the total polyphonic music signal is represented as merged outputs from multiple voice HMMs (Fig.~\ref{fig:MergedOutputHMM}).

Mathematically the model is described as follows.
Let us consider the case of two voices indexed by a variable $s=1,2$, and let $i^{(s)}$ denote the state variable, let $\pi_s(i',i)=P(i|i',s)$ denote the transition probability and let $\phi_s(x;i)=P(x|i,s)$ denote the output probability of each voice HMM (for some output symbol $x$).
For each instance $n$, one voice $s_n$ is chosen by a Bernoulli process as $s_n\sim{\sf Ber}(\alpha_1,\alpha_2)$ where {\sf Ber} is the Bernoulli distribution and its probability parameter $\alpha_{s_n}$ represents how likely the $n$-th output is generated from the HMM of voice $s_n$.
The chosen voice HMM then makes a state transition and outputs $x_n$ while the other voice HMM stays at the current state.
The whole process is described as an HMM with a state space indexed by $k=(s,i^{(1)},i^{(2)})$ and the transition and output probabilities (in the non-interacting case \cite{ICMC}) are given as
\begin{align}
&P(k_n=k|k_{n-1}=k')\notag\\
&\quad= \alpha_s\pi_s(i'^{(s)},i^{(s)})\big(\delta_{s1}\delta_{i'^{(2)}i^{(2)}}+\delta_{s2}\delta_{i'^{(1)}i^{(1)}}\big),\label{eq:TrProbStandardMOHMM}\\
&P(x_n|k_n=k)= \phi_s(x_n;i^{(s)})
\end{align}
where $\delta$ is Kronecker's delta.
A merged-output HMM with more than two voices can be constructed similarly.

As discussed in Ref.~\cite{NakamuraSMC2016}, the merged-output HMM can be seen as a variant of factorial HMM \cite{FactorialHMM} in its most general sense.
Unlike the standard factorial HMM, only one of the voice HMMs makes a state transition and outputs a symbol at each instant.
Owing to this property the sequential regularity within each voice can be described efficiently in the merged-output HMM, even when notes in one voice are interrupted (in the time order) by notes of other voices.
Accordingly necessary inference algorithms are also different as we will see in Sec.~\ref{sec:InferenceAlgorithm}.

%%%%%%%%%%%%%%%%%%%%%%%%%%%%%%%%%%%%%%%%%%%%%%
\section{Proposed Method}
%%%%%%%%%%%%%%%%%%%%%%%%%%%%%%%%%%%%%%%%%%%%%%

We present a complete description of a rhythm transcription method based on merged-output HMM \cite{NakamuraSMC2016} that describes polyphonic performances with multiple-voice structure.
The generative model is presented in Sec.~\ref{sec:ModelFormulation}, the determination of model parameters is discussed in Sec.~\ref{sec:Parameters} and its inference algorithm that simultaneously yields rhythm transcription and voice separation is derived in Sec.~\ref{sec:InferenceAlgorithm}.

\subsection{Model Formulation}\label{sec:ModelFormulation}

A merged-output HMM for rhythm transcription proposed in Ref.~\cite{NakamuraSMC2016} is reviewed here with additional details.
First, the description of the chord model is given, which was explained as a `self-transition' in the note-value state space.
Since self-transition is also used to represent repeated note values of two note clusters, it should be treated with care and we introduce a two-level hierarchical Markov model to solve the problem.
Second, a refinement of switching the probability of choosing voice HMMs is given, which was not mentioned previously but necessary to improve the accuracy of voice separation.

In the following, a music score is specified by multiple sequences, corresponding to voices, of pitches and note values and a MIDI performance signal is specified by a sequence of pitches and onset times.
In this paper we only consider note onsets and thus note length and IOI mean the same thing.

\subsubsection{Model for Each Voice}\label{sec:PartHMM}

A voice HMM is constructed based on the note HMM \cite{Takeda2002}, which is extended to explicitly model pitches in order to appropriately describe voices.
If there are no chords, a score note is specified by a pair of pitch and note value.
Note that to define $N$ note lengths we need $N{+}1$ note onsets and thus $N{+}1$ score notes should be considered.
Let $N{+}1$ be the number of score notes in one voice and let $r_n$ denote the note value of the $n$-th note.
If there are no chords, the note values $\bm r=(r_n)_{n=1}^{N+1}$ are generated by a Markov chain with the following probabilities:
\begin{align}
P(r_1)&=\pi^{\rm ini}(r_1),\\
P(r_{n+1}|r_n)&=\pi(r_n,r_{n+1})\quad(n=1,\ldots,N)
\end{align}
where $\pi^{\rm ini}$ is the initial probability and $\pi$ is the (stationary) transition probability.

\begin{figure}[t]
\begin{center}
{\includegraphics[clip,width=1\columnwidth]{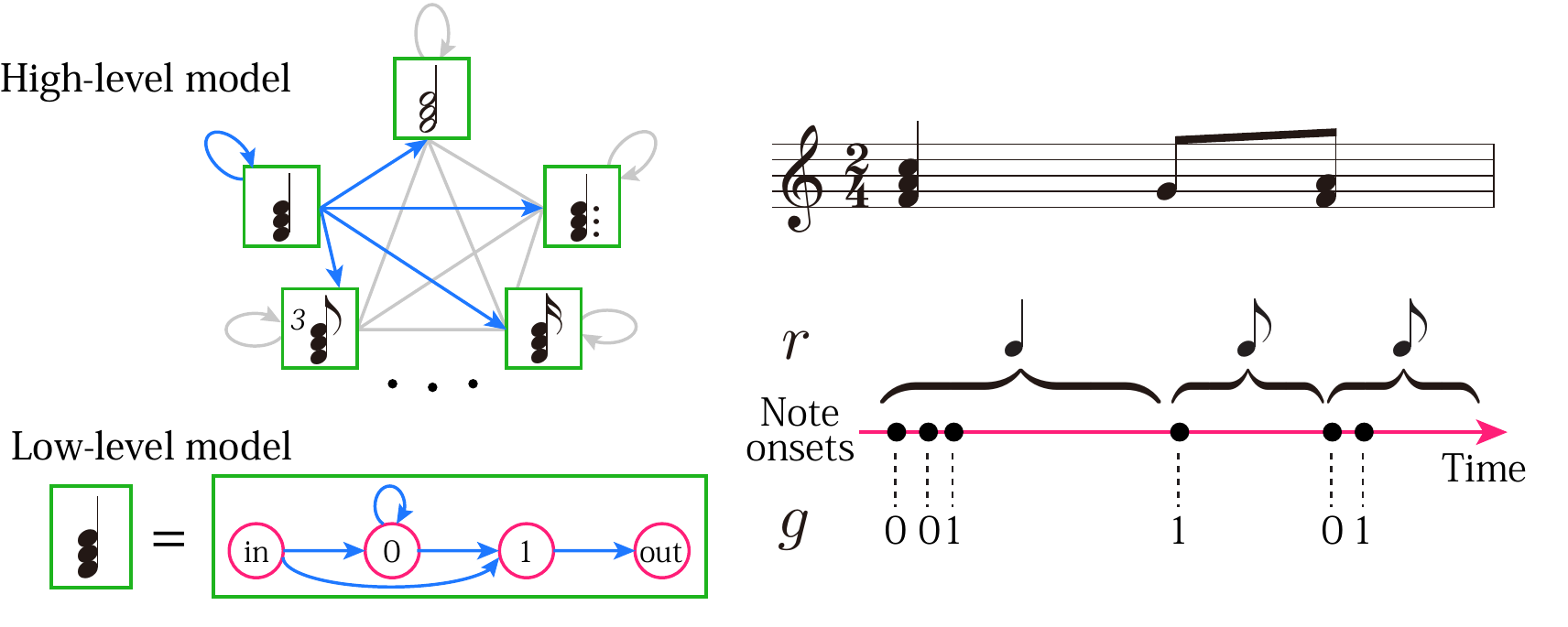}}
\end{center}
%\vspace{-5mm}
\caption{Hierarchical Markov model for sequences of note clusters. Left: The high-level model describes the sequence in units of note clusters and the low-level model describes internal structure of notes in each note cluster. Right: An example polyphonic score and its note onsets represented by the model.}
\label{fig:ChordModel}
%\vspace{-3mm}
\end{figure}
To describe chords, we extend the above Markov model to a two-level hierarchical Markov model with state variables $(r,g)$.
The variable $r$ represents the note value of a note cluster and $g$ indicates whether the next note onset belongs to the same note cluster or not: If $g_n=0$ the $n$-th and $(n{+}1)$-th notes are in a note cluster and if $g_n=1$, they belong to different note clusters.
The variable $g$ also takes the values `in' and `out' to define the initial and exiting probabilities.
The internal Markov model has the topology illustrated in Fig.~\ref{fig:ChordModel} and is described with the following transition probabilities ($\rho^{(r)}_{g,g'}=P(g'|g;r)$):
\begin{align}
\rho^{(r)}_{{\rm in},0}&=\beta_r,\quad \rho^{(r)}_{{\rm in},1}=1-\beta_r,\\
\rho^{(r)}_{0,0}&=\gamma_r,\quad \rho^{(r)}_{0,1}=1-\gamma_r,\quad \rho^{(r)}_{0,{\rm out}}=0,\\
\rho^{(r)}_{1,0}&=\rho^{(r)}_{1,1}=0,\quad \rho^{(r)}_{1,{\rm out}}=1
\end{align}
where $\beta_r$ and $\gamma_r$ are parameters controlling the number of notes in a note cluster.
Denoting $w=(r,g)$ and $w'=(r',g')$, the transition probability of the hierarchical model is given as
\begin{align}
\xi(w',w):=P(w|w')=\rho^{(r')}_{g',{\rm out}}\pi(r',r)\rho^{(r)}_{{\rm in},g}+\delta_{rr'}\rho^{(r)}_{g',g}.
\end{align}
The initial probability is given as $\xi^{\rm ini}(w_1):=P(w_1)=\pi^{\rm ini}(r_1)\rho^{(r_1)}_{{\rm in},g_1}$.
We notate $\bm w=(w_n)_{n=1}^{N{+}1}=(r_n,g_n)_{n=1}^{N+1}$.

To describe the temporal fluctuations, we introduce a tempo variable, denoted by $v_n$, that describes the local (inverse) tempo for the time interval between the $n$-th and $(n{+}1)$-th note onsets \cite{Raphael1999,Nakamura2015}.
To represent the variation of tempos, we put a Gaussian Markov process on the logarithm of the tempo variable $u_n={\rm ln}\,v_n$ as 
\begin{align}
P(u_1)&={\sf N}(u_1;u_{\rm ini},\sigma_{v,{\rm ini}}^2),
\label{eq:TempoIniProb}
\\
P(u_{n+1}|u_n)&={\sf N}(u_{n+1};u_n,\sigma_v^2)
\label{eq:TempoTrProb}
\end{align}
where ${\sf N}(\;\cdot\;;\mu,\Sigma)$ denotes the normal distribution with mean $\mu$ and variance $\Sigma$, and $u_{\rm ini}$, $\sigma_{v,{\rm ini}}$ and $\sigma_v$ are parameters.
The parameter $\sigma_v$ describes the amount of tempo changes.
If the $n$-th and $(n{+}1)$-th notes belong to a note cluster (i.e.\ $g_n=0$), their IOI approximately obeys an exponential distribution \cite{Nakamura2015} and the probability of the onset time of the $(n{+}1)$-th note, denoted by $t_{n+1}$, is then given as
\begin{equation}\label{eq:OnsetTimeProb1}
P(t_{n+1}|t_n,v_n,r_n,g_n=0)={\sf Exp}(t_{n+1};\lambda)
\end{equation}
where {\sf Exp} denotes the exponential distribution and $\lambda$ is the scale parameter, which controls the asynchrony of note onsets in a note cluster.
Otherwise, $t_{n+1}-t_{n}$ has a duration corresponding to note value $r_n$ and the probability is described with a normal distribution as
\begin{equation}\label{eq:OnsetTimeProb2}
P(t_{n+1}|t_n,v_n,r_n,g_n=1)= {\sf N}(t_{n+1};t_n+r_nv_n,\sigma_t^2).
\end{equation}
Intuitively the parameter $\sigma_t$ describes the amount of onset-time fluctuations due to human motor noise when a performer keeps a tempo.
We do not put a distribution on the onset time of the first note $t_1$ because we formulate the model to be invariant under time translations and this value would not affect any results of inference.
We notate $\bm v=(v_n)_{n=1}^{N}$ and $\bm t=(t_n)_{n=1}^{N{+}1}$.

Finally we describe the generation of pitches $\bm p=(p_n)_{n=0}^{N{+}1}$ as a Markov chain (we introduce an auxiliary symbol $p_0$ for later convenience).
The probabilities are
\begin{align}
P(p_1)&=\theta(p_0,p_1),\label{eq:PitchIniProb}\\
P(p_n|p_{n-1})&=\theta(p_{n-1},p_n)\quad(n=2,\ldots,N{+}1)
\end{align}
where $\theta(p',p)$ denotes the (stationary) transition probability and if $p'=p_0$ it denotes the initial probability.

\begin{figure}[t]
\begin{center}
{\includegraphics[clip,width=0.8\columnwidth]{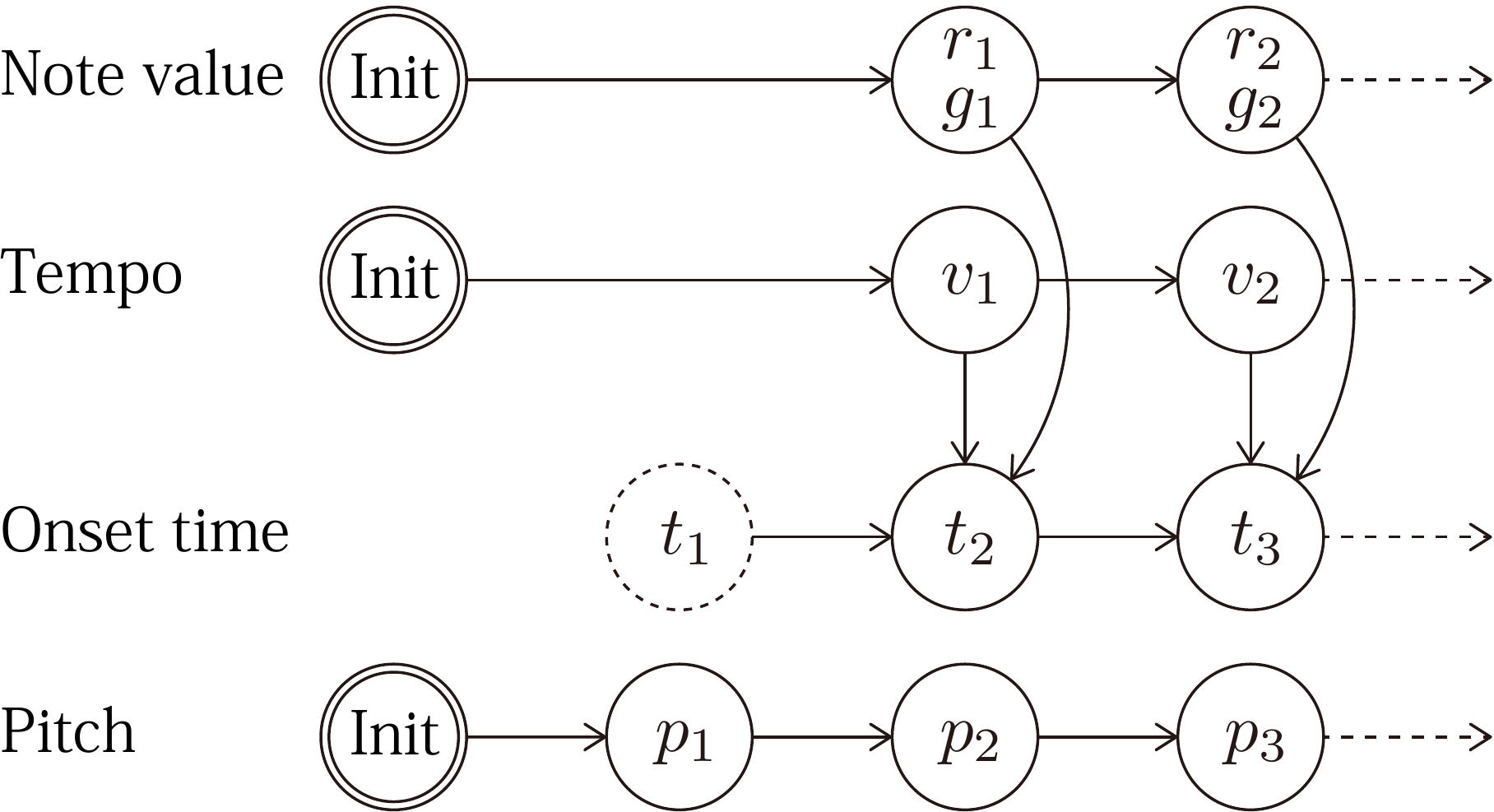}}
\end{center}
%\vspace{-5mm}
\caption{Graphical representation of the autoregressive HMM for one voice. The label `Init' indicates the initial probability and the dotted circle of the first onset time $t_1$ indicates that a distribution is not given for this variable.}
\label{fig:GraphVoiceHMM}
%\vspace{-3mm}
\end{figure}
The above model can be summarised as an autoregressive HMM with hidden states $(\bm r,\bm g,\bm v)$ and outputs $(\bm p,\bm t)$ (Fig.~\ref{fig:GraphVoiceHMM}), which will be a voice HMM.
Although so far the probabilities of pitches are independent of other variables, they will be significant once multiple voice HMMs are merged and the posterior probabilities are inferred.

\subsubsection{Model for Multiple Voices}\label{sec:MultiVoiceModel}

We combine the multiple voice HMMs in Sec.~\ref{sec:PartHMM} using the framework of merged-output HMMs (Sec.~\ref{sec:MOHMM}).
Since in piano performances, which are our main focus, polyrhythm and voice asynchrony usually involve the two hands, we consider a model with two voices, leaving a note that it is not difficult to formalise a model with more than two voices.
In what follows, voices are indexed by a variable $s=1,2$, corresponding to the left and right hand in practice.
All the variables and parameters are now considered for each voice and thus $\bm r^{(s)}=(r^{(s)}_n)_{n=1}^{N_s+1}$ is the sequence of note values in voice $s$, $\pi_s(r',r)$ their transition probability, etc.
Simply speaking, the sequence of merged outputs is obtained by gathering the outputs of the voice HMMs and sorting them according to onset times.
To derive inference algorithms that are computationally tractable, however, we should formulate a model that outputs notes incrementally in the order of observations.
This can be done by introducing stochastic variables $\bm s=(s_n)_{n=1}^{N{+}1}$, which indicate that the $n$-th observed note belongs to voice $s_n$ and follow the probability $s_n\sim{\sf Ber}(\alpha_1,\alpha_2)$.
The parameter $\alpha_{s_n}$ represents how likely the $n$-th note is generated from the HMM of voice $s_n$.

The variable $s_n$ is determined in advance to the pitch, note value or onset time of the corresponding note in the generative process (which is described below).
For rhythm transcription, however, dependence of the parameter $\alpha_s$ on features of the given input (MIDI performance) can be introduced to improve the accuracy of voice separation.
As such a feature, we use contexts of pitch that reflects the constraint on pitch intervals that can be simultaneously played by one hand.
Defining $p_n^{\rm high}$ and $p_n^{\rm low}$ as the highest and lowest pitch that is sounding simultaneously (but not necessarily having a simultaneous onset) with $p_n$, we switch the value of $\alpha_{s_n}$ depending on whether $p_n-p_n^{\rm low}>15$ or not and whether $p_n^{\rm high}-p_n>15$ or not (total of four cases), reflecting the fact that a pitch interval larger than 15 semitones is rarely played with one hand at a time.
The effect of using this context-dependent $\alpha_s$ is examined in Sec.~\ref{sec:ExaminingModel}.

If voice $s_n$ is chosen, then the HMM of voice $s_n$ outputs a note, and the hidden state of the other voice HMM is unchanged.
Such a model can be described with an HMM with a state space labelled by $k=(s,p^{(1)},w^{(1)},t^{(1)},p^{(2)},w^{(2)},t^{(2)},v)$.
Here we have a single tempo variable $v$ that is shared by the two voices in order to assure loose synchrony between them.
The transition probability $P(k_n{=}k|k_{n-1}{=}k')$, for $n\geq2$, is given as
\begin{align}
&\alpha_{s}P(v|v')A_s(w^{(s)},p^{(s)},t^{(s)}|w'^{(s)},p'^{(s)},t'^{(s)};v')
\notag\\
&\!\cdot\Big[\delta_{s1}\delta_{w'^{(2)}w^{(2)}}\delta_{p'^{(2)}p^{(2)}}\delta(t'^{(2)}-t^{(2)})+(1\leftrightarrow2)\Big]
\label{eq:TrProbMergedOutputHMM}
\end{align}
where the expression `$(1{\leftrightarrow}2)$' means that the previous term is repeated with 1 and 2 interchanged and we have defined
\begin{align}
&A_s(w^{(s)},p^{(s)},t^{(s)}|w^{\prime(s)},p^{\prime(s)},t^{\prime(s)};v')
\notag\\
&=\xi_s(w'^{(s)},w^{(s)})\theta_s(p'^{(s)},p^{(s)})P(t^{(s)}|t'^{(s)},v',w'^{(s)})
\end{align}
and $\delta$ denotes Kronecker's delta for discrete variables and Dirac's delta function for continuous variables.
The probability $P(v|v')$ is defined in Eq.~(\ref{eq:TempoTrProb}), and $P(t^{(s)}|t'^{(s)},v',w'^{(s)})$ is defined in Eqs.~(\ref{eq:OnsetTimeProb1}) and (\ref{eq:OnsetTimeProb2}).
For note values the initial probability is given as $P(r_1^{(s)})=\pi_s^{\rm ini}(r_1^{(s)})$, and for pitches the initial probability is given in Eq.~(\ref{eq:PitchIniProb}).
The first onset times $t^{(1)}_1$ and $t^{(2)}_1$ do not have distributions, as explained in Sec.~\ref{sec:PartHMM}, and we practically set $t^{(1)}_1=t^{(2)}_1=t_1$ (the first observed onset time).
Finally the output of the model is given as
\begin{equation}
p_n=p^{(s_n)}_n,\quad t_n=t^{(s_n)}_n,
\end{equation}
and thus the complete-data probability is written as
\begin{equation}
P(\bm k,\bm p,\bm t)=\prod_{n=1}^{N+1}P(k_n|k_{n-1})\delta_{p_np_n^{(s_n)}}\delta(t_n-t_n^{(s_n)}).
\end{equation}
Here $N=N_1+N_2$ denotes the total number of score notes, and the following notations are used: $\bm v=(v_n)_{n=1}^N$, $\bm p=(p_n)_{n=1}^{N+1}$, $\bm t=(t_n)_{n=1}^{N+1}$, and $\bm k=(k_n)_{n=1}^{N{+}1}$.
Note that whereas $\bm p$ and $\bm t$ are observed quantities, $\bm p^{(1)},\bm p^{(2)},\bm t^{(1)}$ and $\bm t^{(2)}$ are not because we cannot directly observe the voice information encrypted in $\bm s$.
\begin{figure}[t]
\begin{center}
{\includegraphics[clip,width=1.\columnwidth]{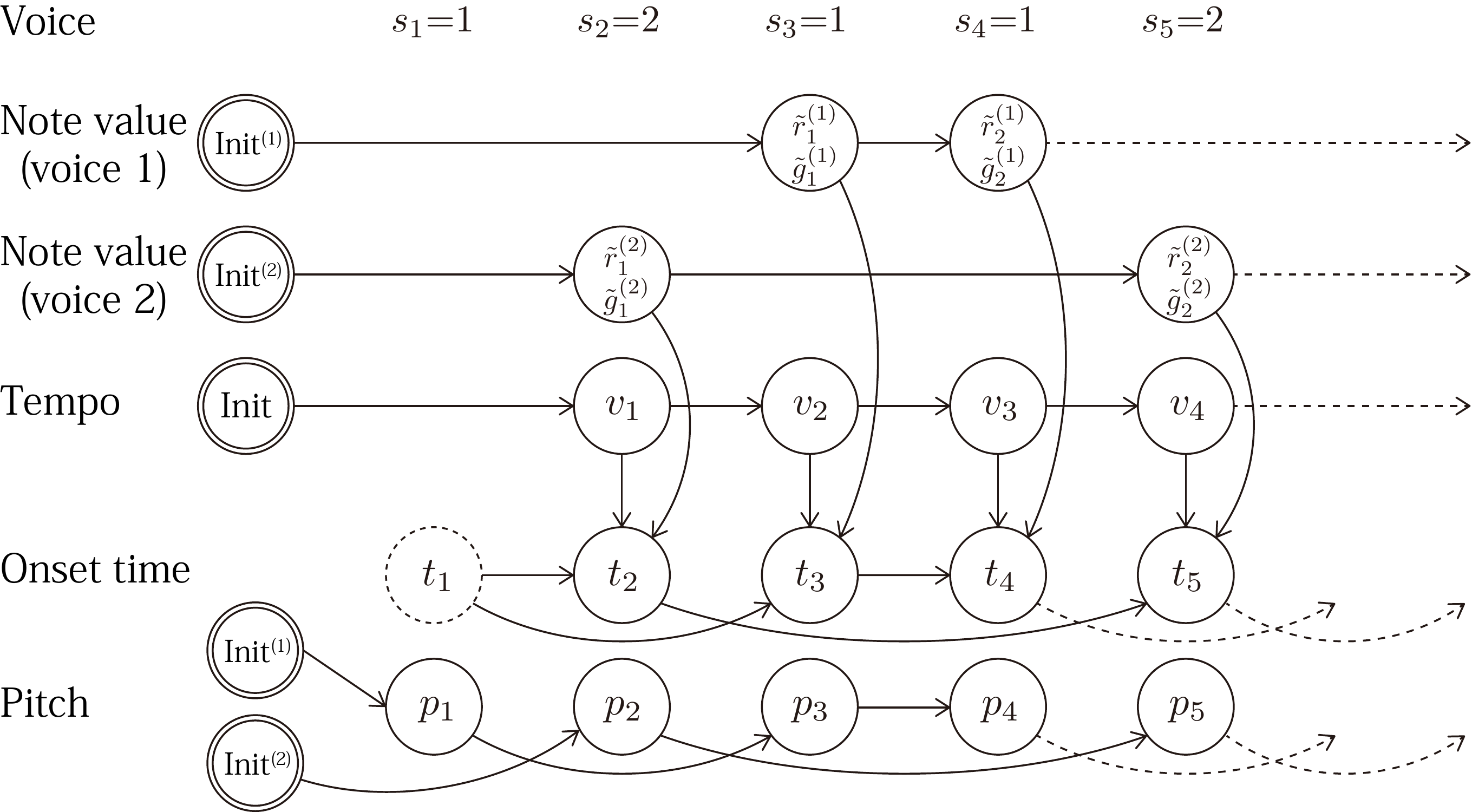}}
\end{center}
%\vspace{-5mm}
\caption{Graphical representation of the proposed merged-output HMM when the voice information is fixed. The variables with a tilde ($\tilde{r}^{(s)}_n$ and $\tilde{g}^{(s)}_n$) represent note values for each voice without redundancies (see Sec.~\ref{sec:InferenceAlgorithm}). See also the caption of Fig.~\ref{fig:GraphVoiceHMM}. Here we have independent initial distributions for note values and pitches of different voice HMMs.}
\label{fig:GraphMOHMM}
\vspace{-3mm}
\end{figure}
The graphical representation of the model is illustrated in Fig.~\ref{fig:GraphMOHMM}.

\subsection{Model Parameters and Their Determination}\label{sec:Parameters}

We here summarise model parameters, explain how they can be determined from data and describe some reasonable constraints to improve the efficiency of parameter learning.
Let $N_p$ and $N_r$ be the number of pitches and note values, which are set as 88 and 15 in our implementation in Sec.~\ref{sec:Evaluation}.

The score model for each voice HMM has the following parameters: $\pi^{\rm ini}(r)$ [$N_r$], $\pi(r,r')$ [$N_r^2$], $\beta_r$ [$N_r$], $\gamma_r$ [$N_r$], $\theta(p_0,p)$ [$N_p$] and $\theta(p,p')$ [$N_p^2$], where the number in square brackets indicate the number of parameters.
(The number of independent parameters may reduce because of normalisation conditions, which we shall not care here for simplicity.)
These parameters can be determined with a data set of music scores with voice indications.
For piano pieces, the two staffs in the grand staff notation can be used for the two voice HMMs.
After representing notes in each voice as a sequence of note clusters as in Fig.~\ref{fig:HomophonizationExample}, $\pi^{\rm ini}(r)$ and $\pi(r,r')$ can be obtained in a standard way.
Determining $\theta(p_0,p)$ and $\theta(p,p')$ is also straightforward.
To determine the parameters $\beta_r$ and $\gamma_r$, we first define the frequency of note clusters containing $m$ notes with note value $r$ as $f^{(r)}_m$.
Since $\beta_r$ is the proportion of note clusters containing more than one notes, it is given by
\begin{equation}
\beta_r=\sum_{m=2}^\infty f^{(r)}_m\bigg/\sum_{m'=1}^\infty f^{(r)}_{m'}.
\end{equation}
The $\gamma_r$ can be obtained by matching the expected staying time at state with $g=0$ (Fig.~\ref{fig:ChordModel}) as follows:
\begin{equation*}
\frac{\sum\limits_{m=2}^\infty mf^{(r)}_m}{\sum\limits_{m'=2}^\infty f^{(r)}_{m'}}
=\langle m\rangle_{m\geq2}
=\sum_{m=2}^\infty m\gamma_r^{m-2}(1-\gamma_r)=\frac{2-\gamma_r}{1-\gamma_r}.
\end{equation*}

In practice, the transition probability $\theta(p,p')$ and the initial probabilities $\theta(p_0,p)$ and $\pi^{\rm ini}(r)$ are often subject to the sparseness problem since the first one has a rather large number of parameters and for the last two only one sample from each piece can be used.
To overcome this problem, we can reduce the number of parameters in the following way, which is used in our implementation.
First, we approximate $\theta(p,p')$ as a function of the interval $p'-p$, which reduces the number of parameters from $N_p^2$ to $2N_p$.
Second, we can approximate $\theta(p_0,p)$ by a gaussian function as
\begin{equation}
\theta(p_0,p)\propto {\sf N}(p;\mu_p,\sigma_p^2).
\end{equation}
Finally, for $\pi^{\rm ini}(r)$, the stationary (unigram) probability obtained from $\pi(r,r')$ can be used.
Note that the pitch probabilities are only used to improve voice separation and their precise values do not much influence the results of rhythm transcription.
Likewise the initial probabilities do not influence the results for most notes due to the Markov property.

The performance model has the following five parameters: $\sigma_v$, $\sigma_v^{\rm ini}$, $v_{\rm ini}(={\rm exp}(u_{\rm ini}))$, $\sigma_t$ and $\lambda$.
These can be determined from performance data, for example, MIDI recordings of piano performances whose notes are matched to the corresponding notes in the scores.
Among these the initial values, $\sigma_v^{\rm ini}$ and $v_{\rm ini}(={\rm exp}(u_{\rm ini}))$ are most difficult to determine from data but again have limited influence as a prior for global tempo, which is supposed to be musically less important (see discussion in Sec.~\ref{sec:EvaluationMethodology}).
In our implementation, they are simply set by hand.
The method for determining the other parameters based on a principle of minimal prediction error is discussed in a previous study \cite{Nakamura2015} and will not be repeated here.

An additional parameter for the merged-output HMM is $\alpha_s$, which is generally obtained by simply counting the number of notes in each voice or can be approximated simply by $\alpha_1=\alpha_2=1/2$.
In our implementation, we obtain four $\alpha_s$'s depending on the context as described in Sec.~\ref{sec:MultiVoiceModel}, which is also straightforward.

\subsection{Inference Algorithm}\label{sec:InferenceAlgorithm}

To obtain the result of rhythm transcription using the model just described, we must estimate the most probable hidden state sequence $\hat{\bm k}={\rm argmax}_{\bm k}P(\bm k|\bm p,\bm t)$ given the observations $(\bm p,\bm t)$.
This gives us the voice information $\hat{\bm s}$ and the estimated note values $\hat{\bm r}^{(1)}$ and $\hat{\bm r}^{(2)}$.
Let $\tilde{\bm w}^{(s)}=(\tilde{w}^{(s)}_n)_{n=1}^{N_s}=(\tilde{r}^{(s)}_n,\tilde{g}^{(s)}_n)_{n=1}^{N_s}$ be the reduced sequence of note values for voice $s$, which is obtained by, for all $n$, deleting the $n$-th element with $s_n\neq s$ in $\hat{\bm w}^{(s)}$.
Then the score time $\tau^{(s)}_n$ of the $n$-th note onset in voice $s$ is given by
\begin{equation}\label{eq:ScoreTime}
\tau^{(s)}_n=\sum_{m=1}^{n-\delta_{ss_1}}\tilde{g}_m^{(s)}\tilde{r}_m^{(s)}.
\end{equation}

The inference algorithm of merged-output HMM has been discussed previously \cite{ISMIR2014}.
Since a merged-output HMM can be seen as an HMM with a product state space, the Viterbi algorithm \cite{Rabiner1989} can be applied for inference in principle.
It was shown that owing to the specific form of transition probability matrix as in Eq.~(\ref{eq:TrProbStandardMOHMM}), the computational complexity for one Viterbi update can be reduced from ${\cal O}(4N_1^2N_2^2)$ to ${\cal O}(2N_1N_2(N_1+N_2))$ where $N_s$ is the size of the state space of the $s$-th voice HMM.
However, since the state space of the model in Sec.~\ref{sec:MultiVoiceModel} involves both discrete and continuous variables, an exact inference in this way is difficult.

To solve this, we discretise the tempo variable, which practically has little influence when the step size is sufficiently small since tempo is restricted in a certain range in conventional music and $v_n$ always has uncertainty of ${\cal O}(\sigma_t/r^{(s_n)}_{n-1})$.
Discretisation of tempo variables has also been used for audio-to-score alignment \cite{Joder2011} and beat tracking \cite{Krebs2015}.
Other continuous variables $\bm t,\bm t^{(1)}$ and $\bm t^{(2)}$ can take only values of observed onset times and thus can, in effect, be treated as discrete variables.
Unfortunately the direct use of the Viterbi algorithm is impractical even with this discretisation.
Let us roughly estimate the computational cost to see this.
Let $N_p$, $N_w$ and $N_v$ be the sizes of the state space for pitch, note value and tempo.
Since onset times $t_n^{(s)}$ could take $N$ values, the size of the state space for $k_n$ is ${\cal O}(2N_p^2N_w^2N^2N_v)$, and the computational cost for one Viterbi update is $\mathscr{C}={\cal O}(4N_p^4N_w^4N^4N_v^2)$.
A rough estimation ($N_p\sim100$, $N_w\sim30$, $N\sim300$, $N_v\sim50$) yields $\mathscr{C}\sim10^{28}$, which is intractable.
Even after using the constraints of the transition probabilities in Eq.~(\ref{eq:TrProbMergedOutputHMM}), we have $\mathscr{C}={\cal O}(4N_p^3N_w^3N^3N_v^2)\sim10^{22}$, which is still intractable.

We can avoid this intractable computational cost and derive an efficient inference algorithm by appropriately relating the hidden variables $(\bm p^{(1)},\bm p^{(2)},\bm t^{(1)},\bm t^{(2)})$ to observed quantities $(\bm p,\bm t)$.
We first introduce a variable $h_n=1,2,\cdots$, which is defined as the smallest $h\geq1$ satisfying $s_n\neq s_{n-h}$ for each $n$.
We find the following relations:
\begin{align}
h_n&=\begin{cases}
h_{n-1}+1, &s_n=s_{n-1};\\
1,&s_n\neq s_{n-1},
\end{cases}
\\
(p^{(s)}_n,t^{(s)}_n)&=\begin{cases}
(p_n,t_n), &s=s_n;\\
(p_{n-h_n},t_{n-h_n}), &s\neq s_n.
\end{cases}
\end{align}
This means that $(\bm p^{(1)},\bm p^{(2)},\bm t^{(1)},\bm t^{(2)})$ are determined if we are given $(\bm s,\bm h,\bm p,\bm t)$ and the effective number of variables is reduced by using $\bm h$.
With this change of variables, we find
\begin{align}
&P(\bm k,\bm p,\bm t)
=P(\bm s,\bm w^{(1)},\bm w^{(2)},\bm v,\bm p^{(1)},\bm p^{(2)},\bm t^{(1)},\bm t^{(2)},\bm p,\bm t)
\notag\\
&=P(\bm s,\bm h,\bm w^{(1)},\bm w^{(2)},\bm v,\bm p,\bm t)
\notag\\
&=\prod_n\bigg\{\alpha_{s_n}P(v_n|v_{n-1})\big[\delta_{s_n1}\delta_{w_n^{(2)}w_{n-1}^{(2)}}+(1\leftrightarrow2)\big]
\notag\\
&\cdot \Big[
\delta_{s_ns_{n-1}}\delta_{h_n(h_{n-1}+1)}A_n^{\rm same}+(1-\delta_{s_ns_{n-1}})\delta_{h_n1}A_n^{\rm diff}\Big]\bigg\},\notag\\
&A_n^{\rm same}=A_{s_n}(w_n^{(s_n)},p_n,t_n|w_{n-1}^{(s_n)},p_{n-1},t_{n-1};v_{n-1}),\notag\\
&A_n^{\rm diff}=A_{s_n}(w_n^{(s_n)},p_n,t_n|w_{n-1}^{(s_n)},p_{\tilde{n}},t_{\tilde{n}};v_{n-1})
\end{align}
where $\tilde{n}$ in the last line should be replaced by $n-h_{n-1}-1$.

We can now apply the Viterbi algorithm on the state space $(\bm s,\bm h,\bm w^{(1)},\bm w^{(2)},\bm v)$.
Noting that the maximum possible value of $h_n$ is $N$ and using the constraints of the transition probabilities, one finds that $\mathscr{C}={\cal O}(4NN_w^3N_v^2)(\sim10^{11})$, which is significantly smaller than the previous values.
Note that so far no ad-hoc approximations have been introduced to reduce the computational complexity.
Practically, we can set a smaller maximal value $N_h(<N)$ of $h_n$ to obtain approximate optimisation, which further reduces the computational cost to ${\cal O}(4N_hN_w^3N_v^2)$.
The number $N_h$ can be regarded as the maximum number of succeeding notes played by one hand without being interrupted by the other hand.
The choice of $N_h$ and its dependency is discussed in Sec.~\ref{sec:DependencyOnNh}.

%%%%%%%%%%%%%%%%%%%%%%%%%%%%%%%%%%%%%%%%%%%%%%
\section{Evaluation}\label{sec:Evaluation}
%%%%%%%%%%%%%%%%%%%%%%%%%%%%%%%%%%%%%%%%%%%%%%

\subsection{Methodology for Systematic Evaluation}\label{sec:EvaluationMethodology}

In a few studies that reported systematic evaluations of rhythm transcription \cite{Otsuki2002,Takeda2002,Cemgil2003}, editing costs (i.e.\ the number of necessary operations to correct an estimated result) are used as evaluation measures.
These studies used the shift operation, which changes the score time of a particular note or equivalently, changes a note value, to count the number of note-wise rhythmic errors.
Musically speaking, on the other hand, the relative note values are more important than the absolute note values, and the tempo error should also be considered.
This is because there is arbitrariness in choosing the unit of note values: For example, a quarter note played in a tempo of 60 BPM has the same duration as a half note played in a tempo of 120 BPM.
Since results of rhythm transcription often contain note values that are uniformly scaled from the correct values, which should not be considered as completely incorrect estimations \cite{Cemgil2000B,Otsuki2002}, we must take into account the scaling operation as well as the shift operation.

\begin{figure}[t]
\begin{center}
{\includegraphics[clip,width=0.8\columnwidth]{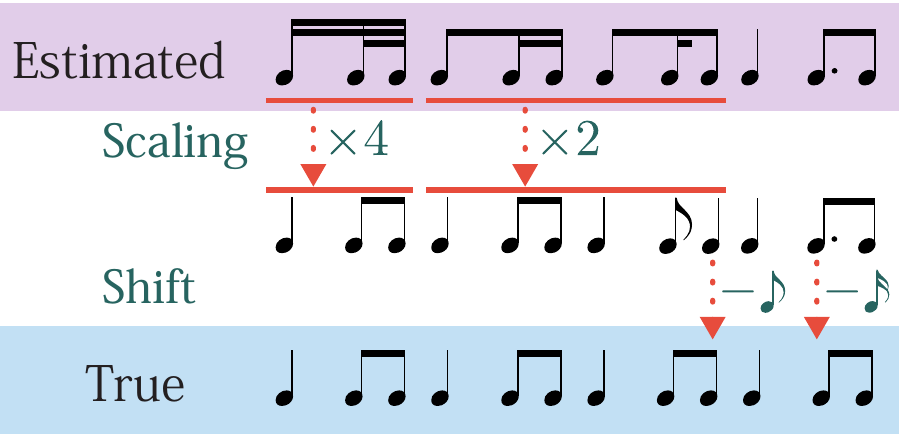}}
\end{center}
%\vspace{-5mm}
\caption{Example of scaling and shift operations to recover the correct transcription from an estimated result.}
\label{fig:RhythmEdition}
%\vspace{-3mm}
\end{figure}
As shown in the example in Fig.~\ref{fig:RhythmEdition}, there can be local scaling operations and shift operations, and a reasonable definition of the editing cost is the least number $N_{\rm o}$ of operations consisting of $N_{\rm sc}$ scaling operations and $N_{\rm sh}$ shift operations ($N_{\rm o}=N_{\rm sc}+N_{\rm sh}$).
As explained in detail in the appendix, this {\it rhythm correction cost} can be calculated by a dynamic programming similarly as the Levenshtein distance.
Definition and calculation of the rhythm correction cost in the polyphonic case are also discussed there.
We use the {\it rhythm correction rate} $\mathscr{R}=N_{\rm o}/N$ as an evaluation measure.

\subsection{Comparisons with Other Methods}\label{sec:Comparison}

We first present results of comparative evaluations.
The purpose is to find out the state-of-the-art method of rhythm transcription and its relation to the proposed model.
Among previous methods described in Sec.~\ref{sec:RelatedWork}, the following six were directly compared: Connectionist Quantizer \cite{Desain1989}, Melisma Analyzers (the first \cite{Temperley1999} and second \cite{Temperley2009} versions), the note HMM \cite{Takeda2002}, the metrical HMM \cite{Raphael2001} and the 2D PCFG model \cite{Tsuchiya2013,Takamune2014}.
The first five are relatively frequently cited and the last one is theoretically important as it provides an alternative way of statistically modelling multiple-voice structure (Sec.~\ref{sec:PolyphonicExtension}).

\subsubsection{Setup}

Two data sets of MIDI recordings of classical piano pieces were used.
One ({\it polyrhythmic} data set) consisted of 30 performances of different (excerpts of) pieces that contained 2 against 3 or 3 against 4 polyrhythmic passages, and the other ({\it non-polyrhythmic} data set) consisted of 30 performances of different (excerpts of) pieces that did not contain polyrhythmic passages.
Pieces by various composers, ranging from J.~S.~Bach to Debussy, were chosen and the players were also various: Some of the performances were taken from the PEDB database \cite{Hashida2008}, a few were performances we recorded, and the rests were taken from collections in public domain websites\footnote{The list of used pieces is available on our web page \cite{DemoPage}.}.

For the proposed method, all normal, dotted, and triplet note values ranging from the whole note to the 32nd note were used as candidate note values.
Parameters for the score model, $\pi(r,r')$, $\beta_r$, $\gamma_r$ and $\theta(p,p')$, and the value of $\alpha_s$ were learned from a data set of classical piano pieces that had no overlap with the test data$^2$.
We set $(\mu_p,\sigma_p)=(54,12)$ for the left-hand voice HMM and $(70,12)$ for the right-hand voice HMM (see Sec.~\ref{sec:Parameters}).
Values for the parameters for the performance model, $\sigma_v$, $\sigma_t$ and $\lambda$, were taken from a previous study \cite{Nakamura2015} (which used performance data different from ours).
The used values were $\bar{\sigma}_v=3.32\times10^{-2}$, $\bar{\sigma}_t=0.02$ s and $\bar{\lambda}=0.0101$ s.
For the tempo variable, we discretised $v_n$ into 50 values logarithmically equally spaced in the range of 0.3 to 1.5 s per quarter note (corresponding to 200 BPM and 40 BPM).
We set $v_{\rm ini}$ as the central value of the range ($0.671$ s per quarter note or 89.4 BPM) and $\sigma_v^{\rm ini}=3\bar{\sigma}_v$.
$N_h$ was chosen as 30, which will be explained later (Sec.~\ref{sec:DependencyOnNh}).

For other methods, we used the default values provided in the source codes, except for the note HMM and the metrical HMM, which are closely related to our method.
For these models, the parameters of the score models were also trained with the same score data set and the performance model was the same as that for the proposed model.
The metrical HMM was build and learned for two cases, duple metres (2/4, 4/4, etc.) and triple metres (3/4, 6/4, etc.), and one of these models were chosen for each performance according to the likelihood.

For Melisma Analyzers, results of the metre analysis were used and the estimated tactus was scaled to a quarter note to use the results as rhythm transcriptions.
For Connectionist Quantizer, which accepts only monophonic inputs, chord clustering was performed beforehand with a threshold of 35 ms on the IOIs of chordal notes.
The algorithm was run for 100 iterations for each performance.
Because this algorithm outputs note lengths in units of 10 ms without indications for tactus, the most frequent note length was taken as the quarter note value.

\subsubsection{Results}\label{sec:Results}

\begin{figure}[t]
\begin{center}
\subfigure[Polyrhythmic data]
{\includegraphics[clip,width=0.9\columnwidth]{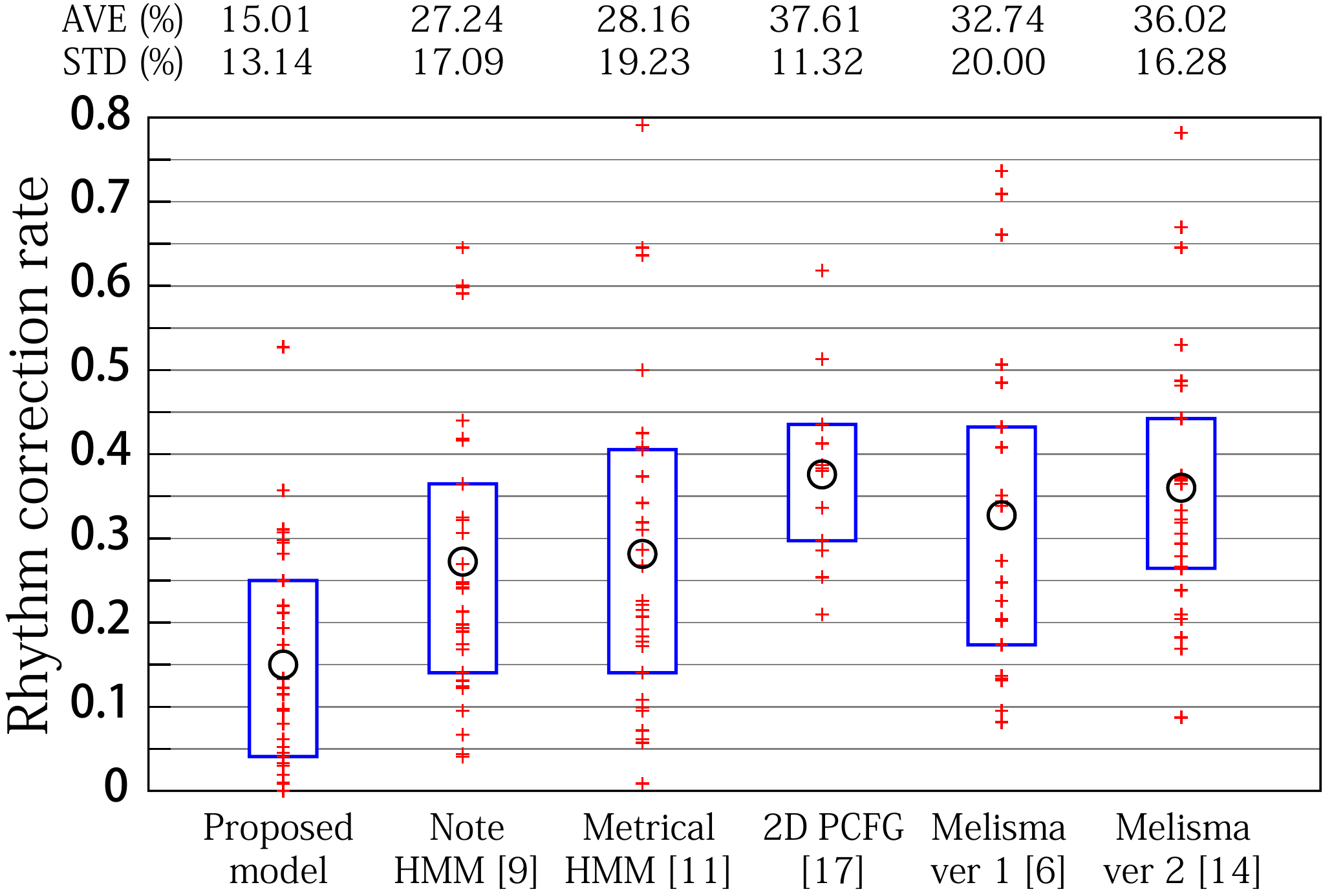}}\\
\subfigure[Non-polyrhythmic data]
{\includegraphics[clip,width=0.9\columnwidth]{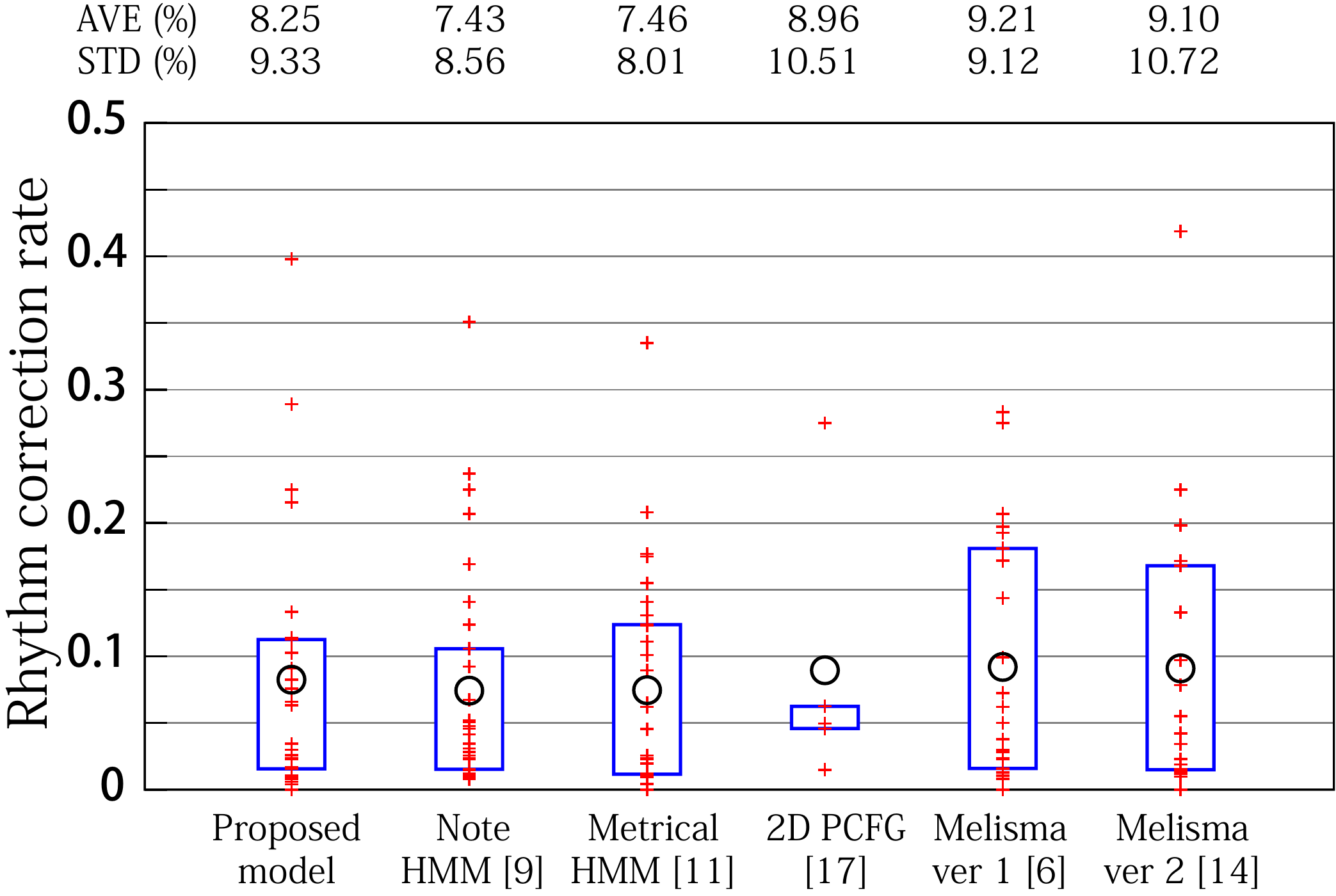}}
\end{center}
\vspace{-3mm}
\caption{Rhythm correction rates (lower is better). The circle indicates the average (AVE), the blue box indicates the range from the first to third quartiles. The standard deviation is indicated as STD.}
\label{fig:RhythmCorrectionRates}
\end{figure}
The distributions of rhythm correction rates, their averages and standard deviations are shown in Fig.~\ref{fig:RhythmCorrectionRates}.
For clear illustration, the results for Connectionist Quantizer, which was much worse than the others, were omitted: The average (standard deviation, first, third quantiles) was 53.7\% (18.5\%, 43.8\%, 67.3\%) for the polyrhythmic data and 38.9\% (13.9\%, 28.2\%, 47.3\%) for the non-polyrhythmic data.

As shown in Table \ref{tab:Result}, some performances were not properly processed by the 2D PCFG model and Melisma Analyzers.
For the 2D PCFG model, because it took much time in processing some performances (executions lasted more than a week for some performances), every performance was run for at least 24 hours and only performances for which execution ended were treated as processed cases.
Among 29 (out of 60) performances for which execution ended, 12 performances did not receive any results (because the parser did not succeed in accepting the performances) and those were also treated as `unprocessed' cases.
To compare the results in the presence of these unprocessed cases, we calculated for each of the algorithms and for successfully processed performances the differences in rhythm corrections rates relative to the proposed model.
Their average and standard error (corresponding to 1$\sigma$ deviation in the $t$-test) are shown in Table \ref{tab:Result}.

\begin{table}[t]
\begin{center}
{\tabcolsep = 2pt
\begin{tabular}{llrr}\toprule
Data set           & Model                            & Unprocessed & $\mathscr{R}-\mathscr{R}_{\rm prop}$ [\%]\\
\midrule
Polyrhythmic  & Note HMM \cite{Takeda2002}       & 0           & ${\bf12.2\pm2.8}$\\
                   & Metrical  HMM \cite{Raphael2001} & 0           & ${\bf13.1\pm3.1}$\\
                   & 2-dim PCFG \cite{Takamune2014}    & 18          & ${\bf23.8\pm3.9}$\\
                   & Melisma ver 1 \cite{Temperley1999} & 9           & ${\bf17.7\pm3.7}$\\
                   & Melisma ver 2 \cite{Temperley2009} & 4           & ${\bf21.4\pm3.3}$\\
                   & Connectionist \cite{Desain1989} & 0           & ${\bf38.7\pm3.2}$\\
\midrule
Non-polyrhythmic & Note HMM \cite{Takeda2002}       & 0           & $-0.82\pm0.50$\\
                   & Metrical  HMM \cite{Raphael2001} & 0           & $-0.79\pm0.61$\\
                   & 2-dim PCFG \cite{Takamune2014}    & 25          & $1.80\pm1.64$\\
                   & Melisma ver 1 \cite{Temperley1999} & 4           & $1.29\pm1.12$\\
                   & Melisma ver 2 \cite{Temperley2009} & 11          & $-0.09\pm1.33$\\
                   & Connectionist \cite{Desain1989} & 0           & ${\bf30.6\pm2.95}$\\
\bottomrule
\end{tabular}
}
\end{center}
%\vspace{-5mm}
\caption{Averages and standard errors of differences of the rhythm correction rate for listed models $\mathscr{R}$ and that of the proposed model $\mathscr{R}_{\rm prop}$ (lower is better). The number of unprocessed pieces (see text) is shown in the third column. Values with a statistical significance $\geq3\sigma$ are illustrated in bold font.}
\label{tab:Result}
\vspace{-5mm}
\end{table}
\begin{figure*}[t]
\begin{center}
{\includegraphics[clip,width=2.\columnwidth]{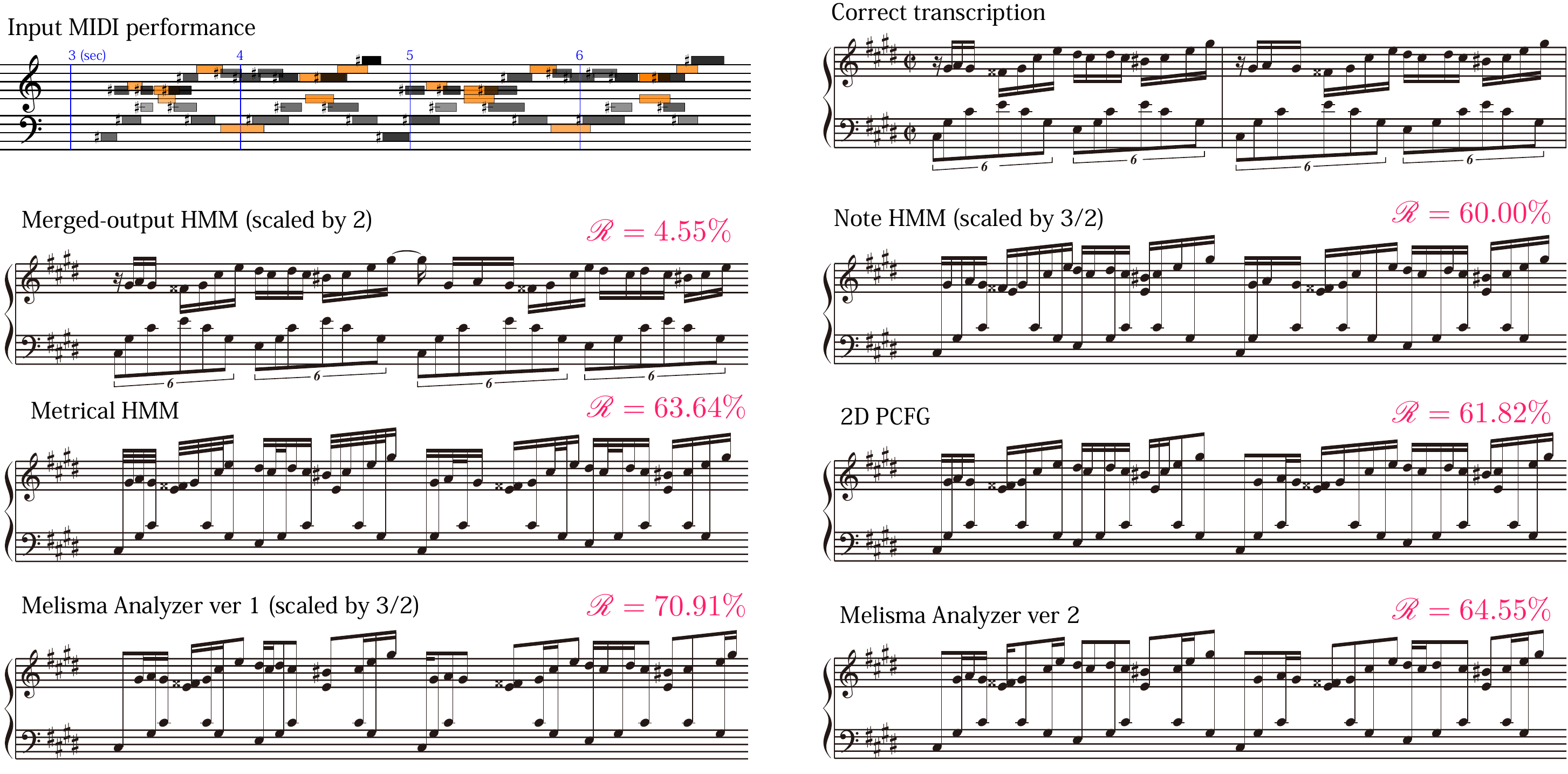}}
\end{center}
\vspace{-5mm}
\caption{Transcription results of a polyrhythmic passage (Chopin's Fantaisie Impromptu). Only part of the performance is illustrated, but the rhythm correction rates for the whole performance are shown. For the result with the proposed model (merged-output HMM), the staffs indicate the estimated voices.}
\label{fig:Example1}
\end{figure*}
\begin{figure*}[t]
\begin{center}
{\includegraphics[clip,width=2.\columnwidth]{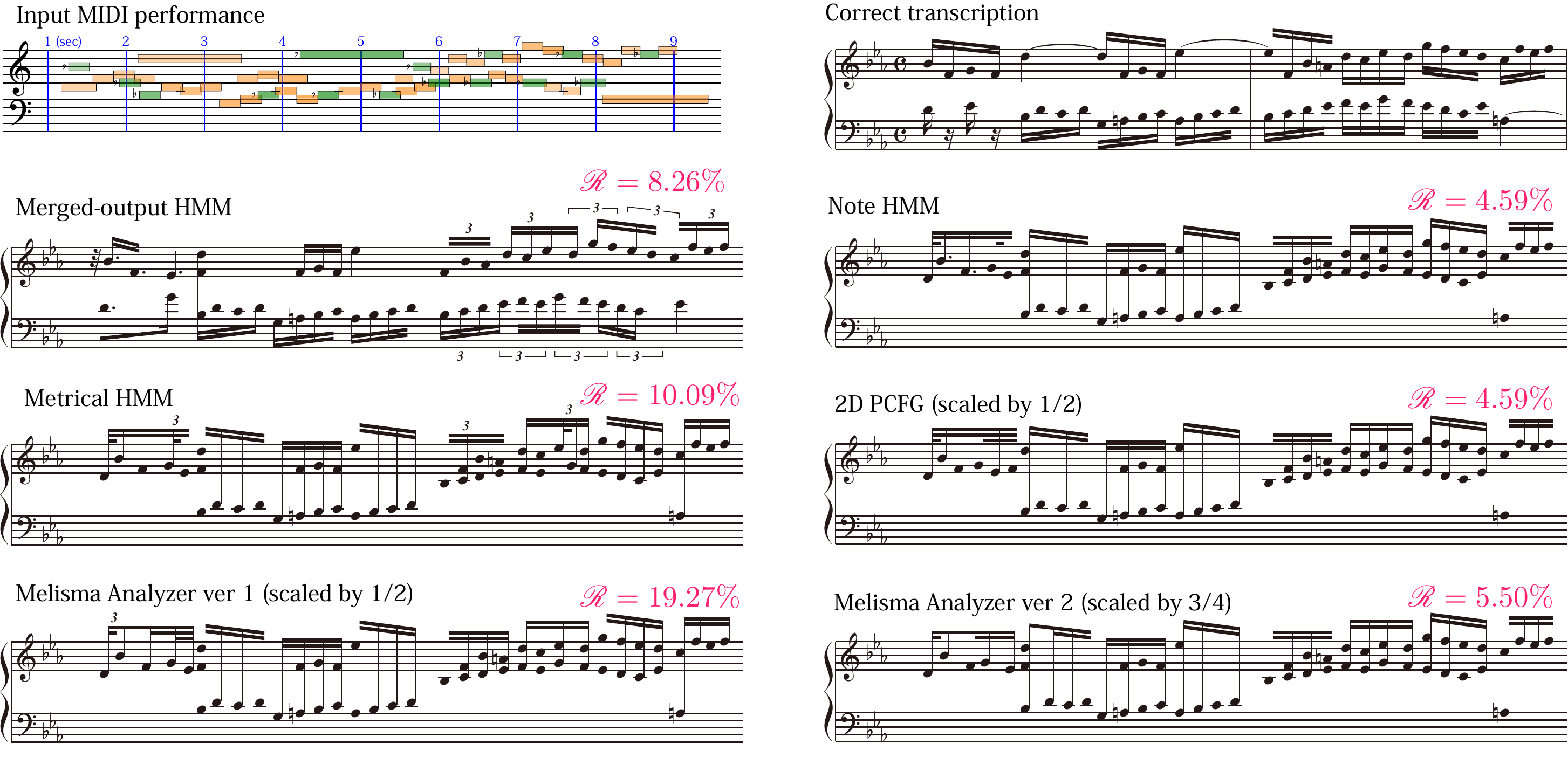}}
\end{center}
\vspace{-5mm}
\caption{Transcription results of a non-polyrhythmic passage (J.~S.~Bach's Invention No.~2). See the caption of Fig.~\ref{fig:Example1}.}
\label{fig:Example2}
\end{figure*}
For the polyrhythmic data, it is clear that the proposed model outperformed the other methods, by more than 12 points in the accuracies and more than 3$\sigma$ deviations in the statistical significances.
Among the other algorithms, the note HMM and metrical HMM had similar accuracies and were second best, and Connectionist Quantizer was the worst.
These results quantitatively confirmed that modelling multiple-voice structure is indeed effective for polyrhythmic performances.
Contrary to our expectation, the result for the 2D PCFG model was second worst for these data.
This might be because that the algorithm using pruning cannot always find the optimal result and the model parameters have not been trained from data, both of which are difficult to ameliorate currently but could possibly be improved in the future.
The results show that the statistical models with (almost fully) learned parameters (the proposed model, the note HMM and the metrical HMM) had better accuracies than the other statistical models with partly learned parameters or without parameter learning (the 2D PCFG model and Melisma Analyzer version 2) and other methods.
A typical example of polyrhythmic performance that is almost correctly recognised by the proposed model but not by other methods is shown in Fig.~\ref{fig:Example1} \footnote{Other examples and sound files are accessible in our demonstration web page \cite{DemoPage}.}.
One finds that the 3 against 4 polyrhythm was properly recognised only by the proposed model (cf.\ Fig.~\ref{fig:HomophonizationExample}).

For the non-polyrhythmic data, Connectionist Quantizer again had a far worst result and the differences among the other methods were much smaller (within 2 points) compared to the polyrhythmic case.
The note HMM and metrical HMM had similar and best accuracies and the proposed model was the third best.
The difference in the average values between the proposed model and the note HMM or the metrical HMM was less than 1 point and the statistical significance was $1.6\sigma$ and $1.3\sigma$, respectively.
Presumably, the main reason that the note and metrical HMMs worked better is that the rhythmic pattern in the reduced sequence of note clusters is often simpler than that of melody/chords in each voice in the non-polyrhythmic case because of the principle of complementary rhythm \cite{SalzerSchachter}.
In particular, notes/chords in a voice can have tied note values that are not contained in our candidate list (e.g.\ a quarter note $+$ 16th note value for the last note of the first bar in the upper staff in the example of Fig.~\ref{fig:Example2}), which can also appear as a result of incorrect voice separation.

It is observed that the transcription by the merged-output HMM can produce desynchronised cumulative note values in different voices (e.g., the quarter note E$\flat$5 in the upper voice in Fig.~\ref{fig:Example2} has a time span different from that of the corresponding notes in the lower voice).
This is due to the lack of constraints to assure the matching of these cumulative note values and the simplification of independent voice HMMs.
For the note HMM and the proposed model, there were grammatically wrong sequences of note values, for example, triplets that appear in single or two notes without completing a unit of beat (e.g.\ the triplet notes in the left-hand part in Fig.~\ref{fig:Example2}).
Further improvements are desired by incorporating such constraints and interactions between voices into the model.

\subsection{Examining the Proposed Method}\label{sec:ExaminingModel}

We here examine the proposed method in more details.

\subsubsection{Dependency on $N_h$}\label{sec:DependencyOnNh}

In Sec.~\ref{sec:InferenceAlgorithm} we introduced a cutoff $N_h$ in the inference algorithm to reduce the computational cost.
In a previous study \cite{ISMIR2014} that discussed the same cutoff, it has been empirically confirmed that $N_h=50$ yields almost exact results for piano performances.
Since it is difficult for our model to run the exact algorithm corresponding to $N_h\to\infty$, we compared results of varying $N_h$ up to 50 to investigate its dependency.

\begin{figure}[t]
\begin{center}
{\includegraphics[clip,width=0.9\columnwidth]{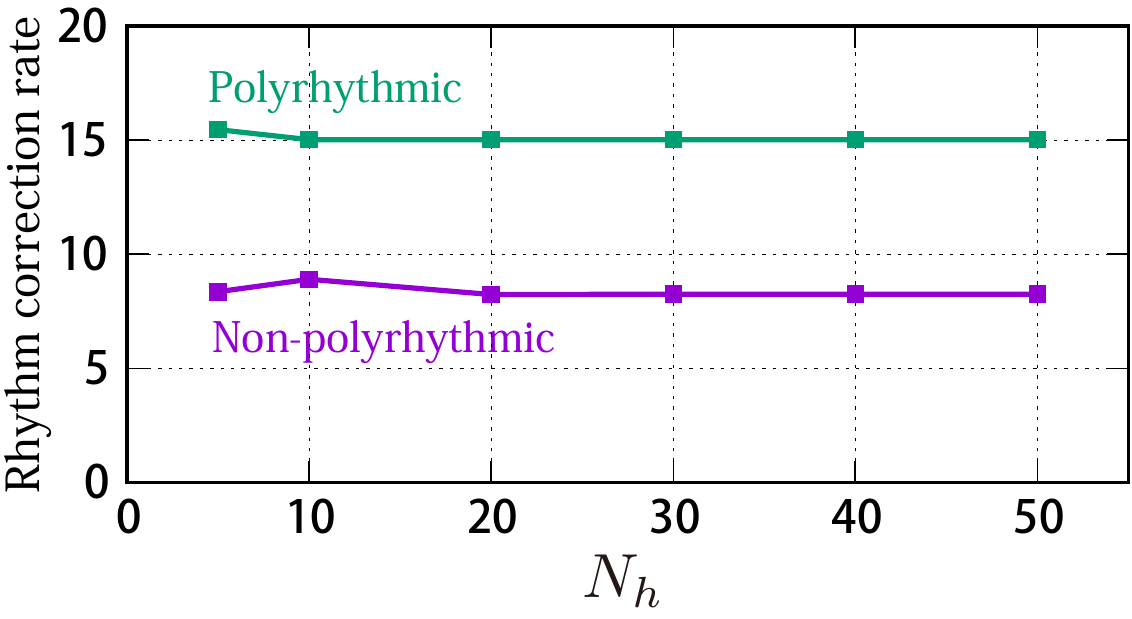}}
\end{center}
\vspace{-5mm}
\caption{Average rhythm correction rates for varying $N_h$.}
\label{fig:VariousNh}
\end{figure}
As shown in Fig.~\ref{fig:VariousNh}, the results were similar for $N_h\geq20$ and were exactly same for $N_h\geq30$.
Based on this result, we used the value $N_h=30$ for all other evaluations in this paper.
Note that the sufficient value of $N_h$ for exact inference may depend on data and that smaller values with sub-optimal estimations could yield better accuracies (as the case of $N_h=20$ for our algorithm and data).

\subsubsection{Effect of the Chord Model}

As explained in Sec.~\ref{sec:ModelFormulation}, we propose a two-level hierarchical HMM for the description of chords, replacing self-transitions used in previous studies \cite{Takeda2007,NakamuraSMC2016}.
To examine its effect in terms of accuracies, we directly compared the two cases implemented in the merged-output HMM.
Since in the former case a self-transition is also used to describe repeated note values of two note clusters, post-processing using the onset-time output probabilities was performed on the results of Viterbi decoding to determine whether a self-transition describe chordal notes or not.

The average rhythm correction rate by the chord model using self-transitions was 15.69\% for the polyrhythmic data and 9.12\% for the non-polyrhythmic data.
By comparing with the values in Fig.~\ref{fig:RhythmCorrectionRates}, our chord model was slightly better and the differences are $0.87\pm0.46$ and $0.68\pm2.47$ (statistical significance $1.9\sigma$ and $0.3\sigma$) for the two data sets.
These results indicate that our chord model is not only conceptually simple but also seems to improve the accuracy slightly.

\subsubsection{Effect of Joint Estimation of Voice and Note Values and Voice Separation Accuracy}

A feature of our method is the simultaneous estimation of voice and note values.
An alternative approach is to use a cascading algorithm that performs voice separation first and then estimates note values using the estimated voices.
To examine the effectiveness of the joint estimation approach, we implemented a cascading algorithm consisting of voice separation using only the pitch part of the model in Sec.~\ref{sec:MultiVoiceModel} and rhythm transcription using two note HMMs with coupled tempos and compared it with the proposed method.

The average rhythm correction rate by the cascading algorithm was 16.89\% for the polyrhythmic data and 9.67\% for the non-polyrhythmic data.
By comparing with the values in Fig.~\ref{fig:RhythmCorrectionRates}, we see that the proposed method was slightly better and the differences are $1.88\pm1.64$ and $1.42\pm0.47$ (statistical significance $1.1\sigma$ and $3.0\sigma$) for the two data sets.
These results indicate the effectiveness of the joint estimation approach of the proposed method while the cascading algorithm may have practical importance because of its smaller computational cost.

We also measured the accuracy of voice separation (into two hands).
The accuracy with the proposed model was 94.2\% for the polyrhythmic data and 88.0\% for the non-polyrhythmic data and with the cascading algorithm it was 93.8\% and 92.5\%.
This indicates firstly that a similar (or higher) accuracy can be obtained by using only the pitch information and secondly that a higher accuracy of voice separation does not necessarily lead to a better rhythm recognition accuracy.

\subsubsection{Influence of the Model Parameters}

In our implementation, parameters of the tempo variables (mainly $\sigma_v$, $\sigma_t$ and $\lambda$) were not optimised but adjusted to values measured in a completely different experiment \cite{Nakamura2015}.
Since these parameters play important roles of describing `naturalness' of temporal fluctuations in music performance, we performed experiments to examine their influence.

\begin{figure}[t]
\begin{center}
\subfigure
{\includegraphics[clip,width=0.9\columnwidth]{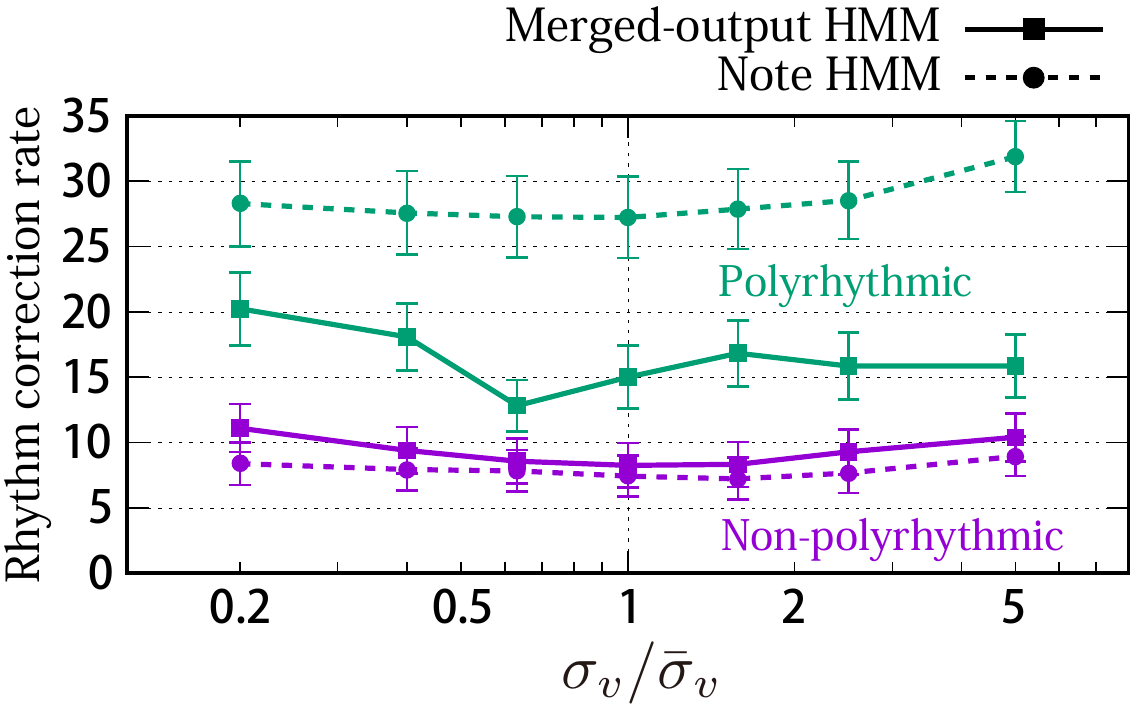}}\\
\vspace{-1mm}
\subfigure
{\includegraphics[clip,width=0.9\columnwidth]{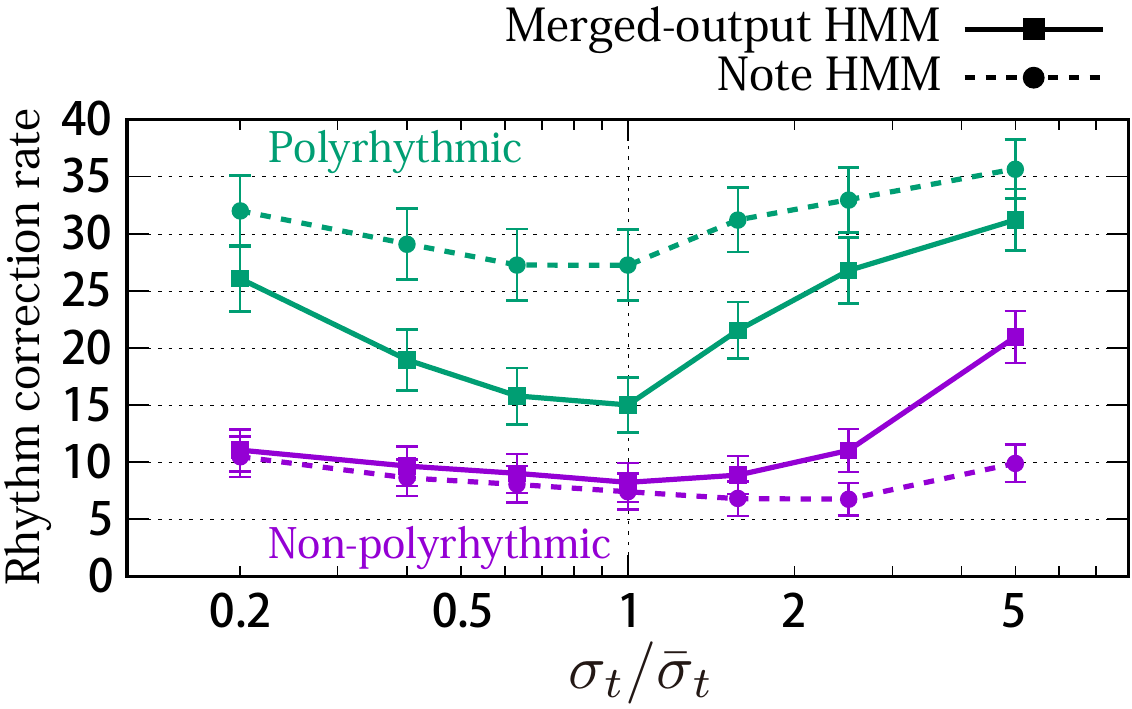}}\\
\vspace{-1mm}
\subfigure
{\includegraphics[clip,width=0.9\columnwidth]{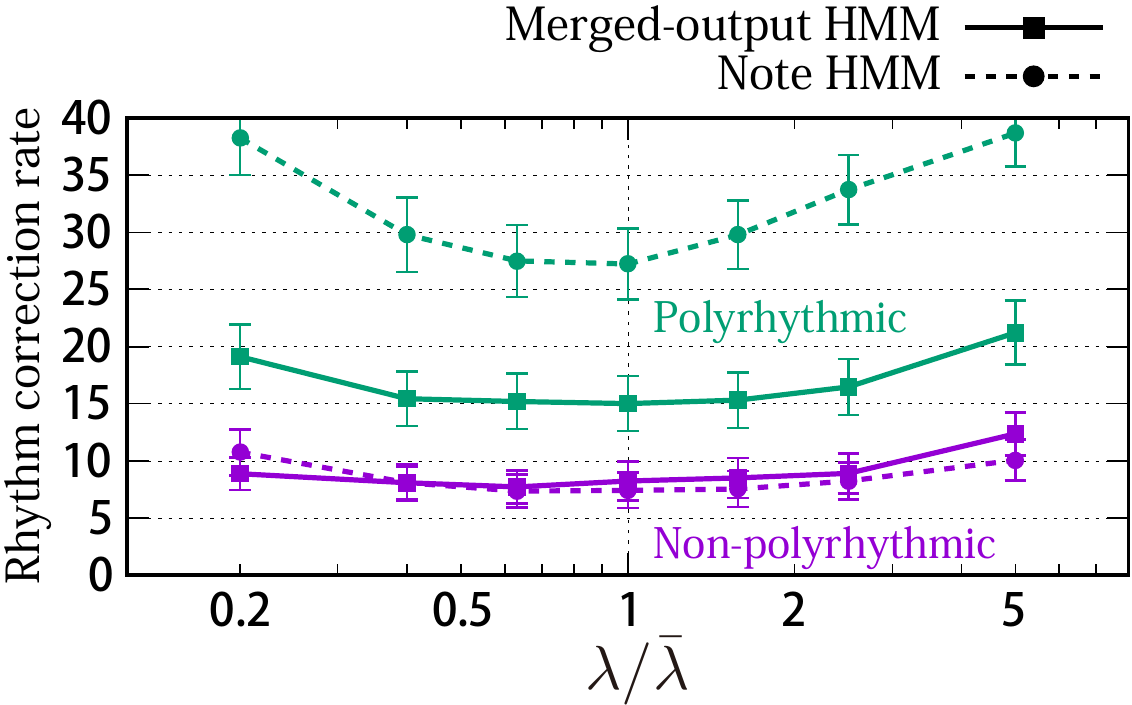}}
\end{center}
\vspace{-3mm}
\caption{Average rhythm correction rates for the merged-output HMM and the note HMM for varying $\sigma_v$, $\sigma_t$ and $\lambda$. Error bars indicate standard errors.}
\label{fig:Influence}
\end{figure}
Fig.~\ref{fig:Influence} shows the results of measuring average rhythm correction rates for varying $\sigma_v$, $\sigma_t$ and $\lambda$ around the value used for our implementation.
When one parameter was varied, the other parameters were fixed to the original values.
Results for the note HMM are also shown as references.
We see that overall (with some exceptions) the parameters were optimal around the original values, which implies the universality of these parameters.
For both models, we found values with a better accuracy (at least for one of the data sets) than the original values, suggesting the possibility of further optimisations.

We see relatively large influence of $\sigma_t$ on the merged-output HMM and $\lambda$ on the note HMM.
This can be explained by the fact that compared to the note HMM, the merged-output HMM must handle a more number of inter-note-cluster durations and a less number of chordal notes because of the presence of two voices.
Accordingly the $\sigma_t$, which controls the fluctuation of inter-note-cluster durations, has more chances and the $\lambda$, which controls the asynchrony of chordal notes, has less chances to influence the results of the merged-output HMM.

Finally we examined the effect of context-dependent $\alpha_s$ described in Sec.~\ref{sec:MultiVoiceModel}.
For this purpose we simply run the proposed method with uniformly distributed $\alpha_s$ ($\alpha_1=\alpha_2=1/2$).
The average rhythm correction rate was slightly worse ($17.96\pm2.49$) for the polyrhythmic data and slightly better ($7.76\pm1.33$) for the non-polyrhythmic data.
On the other hand, the accuracy of voice separation was 30.4\% (50.0\%) for the polyrhythmic (non-polyrhythmic) data, which is much worse.
The results confirm that the context-dependent $\alpha_s$ is important to improve voice separation and provide yet another example that a more precise voice separation does not necessarily induce better rhythm recognition accuracy.

%%%%%%%%%%%%%%%%%%%%%%%%%%%%%%%%%%%%%%%%%%%%%%
\section{Conclusion}
%%%%%%%%%%%%%%%%%%%%%%%%%%%%%%%%%%%%%%%%%%%%%%

We have described and examined a rhythm transcription method based on a merged-output HMM of polyphonic symbolic performance.
This model has an internal structure consisting of multiple HMMs to solve the long-standing problem of properly describing the multiple-voice structure of polyphonic music.
With the inference method derived in this paper, the algorithm can perform voice separation and note-value recognition simultaneously.
The technique of deriving inference algorithms with reduced computational cost can be applied to other merged-output HMMs with autoregressive voice HMMs, which are expectedly effective models of polyphonic music where the multiple-voice structure is significant.

By examining the proposed method, we also confirmed that simultaneously inferring the voice and rhythm information improved the accuracy of rhythm transcription compared to a cascading approach, even though it did not necessarily improve the accuracy of voice separation. 
On the other hand, transcribed results sometimes contained unwanted asynchrony between notes in different voices that have almost simultaneous notes onset times.
This is because the model describes no information about the absolute onset time and there are no strong interactions between voices other than the shared tempo.
The use of merged-output HMMs with interacting voice HMMs \cite{ICMC} could provide a solution in principle, but how to describe synchrony of global score times while retaining computational tractability is a remaining problem.

With evaluations comparing seven rhythm transcription methods, we found that the proposed method performed significantly better than others for polyrhythmic performances.
For non-polyrhythmic performances, we found that the note HMM and metrical HMM had the best accuracies and the proposed method was almost as good as (but slightly worse than) these methods.
These results revealed the state-of-the-art methods for rhythm transcription that were different for the two kinds of data.
While practically running two or more methods simultaneously and choosing the best result can be effective, developing a unified method that yields best results for both kinds of data is desired.
Solving the above problem of unwanted asynchrony would be one key and constructing a model with variable number of voices would be another.

\appendices

%%%%%%%%%%%%%%%%%%%%%%%%%%%%%%%%%%%%%%%%%%%%%%
\section{Calculation of the Rhythm Correction Cost}
%%%%%%%%%%%%%%%%%%%%%%%%%%%%%%%%%%%%%%%%%%%%%%

Let us formulate the rhythm correction cost introduced in Sec.~\ref{sec:EvaluationMethodology} and derive an algorithm for calculating it.
We first consider the monophonic case.
Let $r_n^{\rm true}$ and $r_n^{\rm est}$ be the correct and estimated note value of the $n$-th note length in the performance input ($n=1,\ldots,N$).
We consider a sequence of pairs $(d_n,e_n)$ of scaling factor $d_n$ and shift interval $e_n$ for $n=1,\ldots,N$.
To recover $(r_n^{\rm true})_{n=1}^N$ from $(r_n^{\rm est})_{n=1}^N$ with the scaling and shift operations, we must have
\begin{equation}\label{eq:noteValueEdition}
r_{n}^{\rm true}=d_nr_{n}^{\rm est}+e_n
\end{equation}
for $n=1,\cdots,N$.
The number of scaling operations and that of shift operations are formally defined as $N_{\rm sc}=\#\{n|d_{n+1}\neq d_n\}$ and $N_{\rm sh}=\#\{n|e_n\neq0\}$.
The minimum number of editing operations $N_{\rm o}$ is determined by minimising $N_{\rm sc}+N_{\rm sh}$ for all sequences $(d_n,e_n)_{n=1}^{N}$ satisfying Eq.~(\ref{eq:noteValueEdition}).
This is a special case of a generalised rhythm correction cost, which can be defined similarly as the minimum of $w_{\rm sc}N_{\rm sc}+w_{\rm sh}N_{\rm sh}$ for some non-negative real numbers $w_{\rm sc}$ and $w_{\rm sh}$.

Let us now present a dynamic programming to calculate the rhythm correction cost $N_{\rm o}$.
We describe a general algorithm valid for any values of $w_{\rm sc}$ and $w_{\rm sh}$.
The algorithm can be derived in the same form as the Viterbi algorithm for HMMs.
We define the `state space' $\Omega$ as the set of all possible scaling operations, which can be constructed by taking ratios of all possible note values.
The space $\Omega$ is finite since the set of note values is finite.
The scaling cost (analogous to transition probability) $C_{\rm sc}:\Omega\times \Omega\to\mathbb{R}$ is defined as
\begin{equation}
C_{\rm sc}(d_{n-1},d_n)=\begin{cases}
0,&{\rm if}~d_n=d_{n-1};\\
w_{\rm sc},&{\rm otherwise}.
\end{cases}
\end{equation}
For the initial value $d_1$ the cost is defined as $C_{\rm sc}(d_1)=0$ if $d_1=1$ and $w_{\rm sc}$ otherwise.
To describe whether a shift operation is necessary for the $n$-th note value after scaling, the shift cost (analogous to output probability) $C_{\rm sh}:\Omega\to\mathbb{R}$ is defined as
\begin{equation}
C_{\rm sh}(d_n)=\begin{cases}
0,&{\rm if}~r_{n}^{\rm true}=d_nr_{n}^{\rm est};\\
w_{\rm sh},&{\rm otherwise}.\end{cases}
\end{equation}
Defining the total cost as
\begin{equation}
C(d_1,\ldots,d_N)=\sum_{n=1}^{N}\left(C_{\rm sc}(d_{n-1},d_n)+C_{\rm sh}(d_n)\right)
\end{equation}
(we understand $C_{\rm sc}(d_0,d_1)$ as $C_{\rm sc}(d_1)$), we have the relation
\begin{equation}\label{eq:RhythmCorrectionCostRelation}
N_{\rm o}=\min_{d_1,\ldots,d_N}C(d_1,\ldots,d_N).
\end{equation}
The right-hand side of Eq.(\ref{eq:RhythmCorrectionCostRelation}) can be calculated by the Viterbi algorithm \cite{Rabiner1989} with computational complexity ${\cal O}((\#\Omega)^2N)<{\cal O}(N_r^4N)$ where $N_r$ is the number of note-value types.

Note that the above formulation is already valid in the presence of chords.
Chordal notes are represented as notes with $r_n=0$.
The error in clustering a chord, i.e., $r_n^{\rm true}=0$ but $r_n^{\rm est}\neq0$ or vice versa, can be corrected by a shift operation.

When there are separated multiple voices, we can apply shift operations on each note in each voice and scaling operations on all voices simultaneously.
If the estimated score time duration between the first notes of any two voices is different from that in the correct score, it must be corrected as well.
The rhythm correction cost for multiple voices can be calculated by the same manner as above using as $(r^{\rm est}_n)_{n=1}^N$ a sequence of by merging all $\tau_{n+1}^{(s)}-\tau_n^{(s)}$ for all $s$ and $1\leq n\leq N_s$ and $\tau_1^{(s)}$ for all $s>1$ in the order of onset time, where $\tau_n^{(s)}$ is the score time of $n$-th note onset in voice $s$ as defined in Eq.~(\ref{eq:ScoreTime}).

%%%%%%%%%%%%%%%%%%%%%%%%%%%%%%%%%%%%%%%%%%%%%%
\section*{Acknowledgment}
%%%%%%%%%%%%%%%%%%%%%%%%%%%%%%%%%%%%%%%%%%%%%%

We are grateful to David Temperley, Norihiro Takamune and Henkjan Honing for providing their source codes.
The author EN\ thanks Hiroaki Tanaka for useful discussions on merged-output HMM and Yoshiaki Bando for his help with running computer programs.
This work is partially supported by JSPS KAKENHI Nos.\ 24220006, 26240025, 26280089, 26700020, 15K16054, 16H01744 and 16J05486, and JST CrestMuse, OngaCREST and OngaACCEL projects.

\ifCLASSOPTIONcaptionsoff
  \newpage
\fi


%%%%%%%%%%%%%%%%%%%%%%%%%%%%%%%%%%%%%%%%%%%%%%
%%%%%%%%%%%%%%%%%%%%%%%%%%%%%%%%%%%%%%%%%%%%%%

%%%%%%%%%%%%%%%%%%%%%%%%%%%%%%%%%%%%%%%%%%%%%%
%%%%%%%%%%%%%%%%%%%%%%%%%%%%%%%%%%%%%%%%%%%%%%


%\end{IEEEbiography}

\begin{IEEEbiographynophoto}{Eita Nakamura}
He received a Ph.D.\ degree in physics from the University of Tokyo in 2012. After having been a post-doctoral researcher at the National Institute of Informatics, Meiji University and Kyoto University, he is currently a JSPS Research Fellow in the Speech and Audio Processing Group at Kyoto University. His research interests include music modelling and analysis, music information processing and statistical machine learning.
\end{IEEEbiographynophoto}

\begin{IEEEbiography}[{\includegraphics[width=1in,height=1.25in,clip,keepaspectratio]{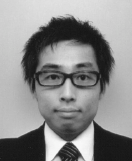}}]{Kazuyoshi Yoshii}
He received the Ph.D. degree in informatics from Kyoto University, Japan, in 2008. He is currently a Senior Lecturer at Kyoto University. His research interests include music signal processing and machine learning. He has received several awards including the IPSJ Yamashita SIG Research Award and the Best-in-Class Award of MIREX 2005. He is a Member of the Information Processing Society of Japan and Institute of Electronics, Information, and Communication Engineers.
\end{IEEEbiography}

\begin{IEEEbiography}[{\includegraphics[width=1in,height=1.25in,clip,keepaspectratio]{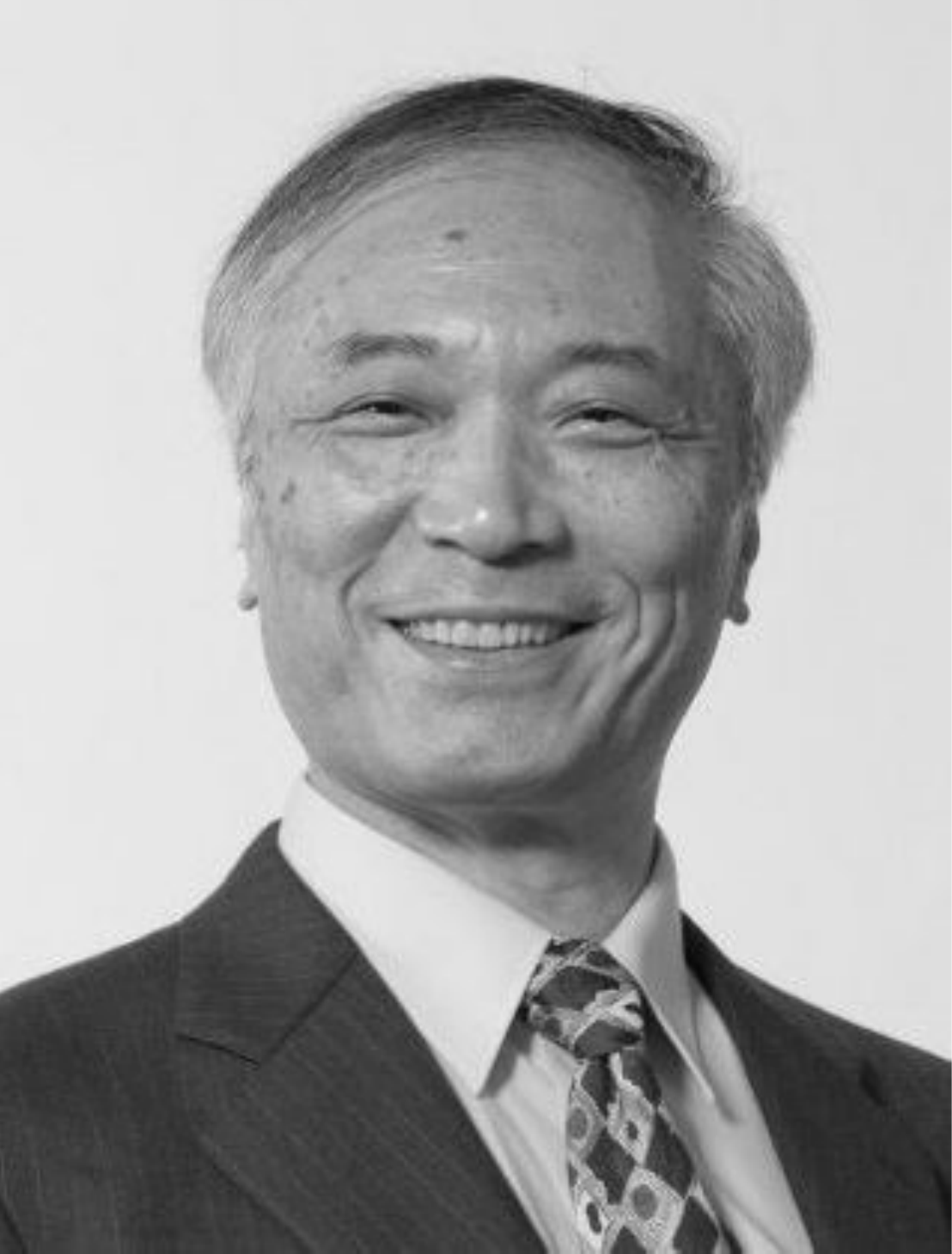}}]{Shigeki Sagayama}
He received the B.E., M.S., and Ph.D. degrees from the University of Tokyo, Tokyo, Japan, in 1972, 1974, and 1998, respectively, all in mathematical engineering and information physics. He joined Nippon Telegraph and Telephone Public Corporation (currently, NTT) in 1974 and started his career in speech analysis, synthesis, and recognition at NTT Labs in Musashino, Japan. From 1990, he was Head of the Speech Processing Department, ATR Interpreting Telephony Laboratories, Kyoto, Japan where he was in charge of an automatic speech translation project. In 1993, he was responsible for speech recognition, synthesis, and dialog systems at NTT Human Interface Laboratories, Yokosuka, Japan. In 1998, he became a Professor of the Graduate School of Information Science, Japan Advanced Institute of Science and Technology (JAIST), Ishikawa. In 2000, he was appointed Professor at the Graduate School of Information Science and Technology (formerly, Graduate School of Engineering), the University of Tokyo. After his retirement from the University of Tokyo, he is a Professor of Meiji University from 2014. His major research interests include the processing and recognition of speech, music, acoustic signals, handwriting, and images. He was the leader of anthropomorphic spoken dialog agent project (Galatea Project) from 2000 to 2003. Prof. Sagayama received the National Invention Award from the Institute of Invention of Japan in 1991, the Director General's Award for Research Achievement from the Science and Technology Agency of Japan in 1996, and other academic awards including Paper Awards from the Institute of Electronics, Information and Communications Engineers, Japan (IEICEJ) in 1996 and from the Information Processing Society of Japan (IPSJ) in 1995. He is a member of the Acoustical Society of Japan, IEICEJ, and IPSJ.
\end{IEEEbiography}


\begin{thebibliography}{99}%\setlength\itemsep{-2pt}

\bibitem{Klapuri2006}
A.~Klapuri and M.~Davy (eds.),
{\it Signal Processing Methods for Music Transcription}, Springer, 2006.

\bibitem{Benetos2013}
E.~Benetos, S.~Dixon, D.~Giannoulis, H.~Kirchhoff and A.~Klapuri,
%E.~Benetos {\it et al.},
``Automatic Music Transcription: Challenges and Future Directions,''
{\it J. Intelligent Information Systems}, vol.~41, no.~3, pp.~407--434, 2013.

\bibitem{LonguetHiggins1987}
H.~Longuet-Higgins,
{\it Mental Processes: Studies in Cognitive Science}, MIT Press, 1987.

\bibitem{Pressing1993}
J.~Pressing and P.~Lawrence,
``Transcribe: A Comprehensive Autotranscription Program,''
{\it Proc.\ ICMC}, pp.~343--345, 1993.

\bibitem{Chowning1984}
J.~Chowning, L.~Rush, B.~Mont-Reynaud, C.~Chafe, W.~Schloss and J.~Smith,
%J.~Chowning {\it et al.},
``Intelligent Systems for the Analysis of Digitized Acoustic Signals,''
{\it Tech.\ Rep.\ CCRMA}, STAN-M-15, 1984.

\bibitem{Temperley1999}
D.~Temperley and D.~Sleator,
``Modeling Meter and Harmony: A Preference-Rule Approach,''
{\it Comp.\ Mus.\ J.}, vol.~23, no.~1, pp.~10--27, 1999.

\bibitem{Desain1989}
P.~Desain and H.~Honing,
``The Quantization of Musical Time: A Connectionist Approach,''
{\it Comp.\ Mus.\ J.}, vol.~13, no.~3, pp.~56--66, 1989.

\bibitem{Otsuki2002}
T.~Otsuki, N.~Saitou, M.~Nakai, H.~Shimodaira and S.~Sagayama,
%T.~Otsuki {\it et al.},
``Musical Rhythm Recognition Using Hidden Markov Model (in Japanese),''
{\it J.\ Information Processing Society of Japan}, vol.~43, no.~2, pp.~245--255, 2002.

\bibitem{Takeda2002}
H.~Takeda, T.~Otsuki, N.~Saito, M.~Nakai, H.~Shimodaira and S.~Sagayama,
%H.~Takeda {\it et al.},
``Hidden Markov Model for Automatic Transcription of MIDI Signals,''
{\it Proc.\ MMSP}, pp.~428--431, 2002.

\bibitem{Takeda2007}
H.~Takeda, T.~Nishimoto and S.~Sagayama,
%H.~Takeda {\it et al.},
``Rhythm and Tempo Analysis Toward Automatic Music Transcription,''
{\it Proc.\ ICASSP}, vol.~4, pp.~1317--1320, 2007.

\bibitem{Raphael2001}
C.~Raphael,
``Automated Rhythm Transcription,''
{\it Proc.\ ISMIR}, pp.~99--107, 2001.

\bibitem{Hamanaka2003}
M.~Hamanaka, M.~Goto, H.~Asoh and N.~Otsu,
%M.~Hamanaka {\it et al.},
``A Learning-Based Quantization: Unsupervised Estimation of the Model Parameters,''
{\it Proc.\ ICMC}, pp.~369--372, 2003.

\bibitem{Cemgil2003}
A.~Cemgil and B.~Kappen,
``Monte Carlo Methods for Tempo Tracking and Rhythm Quantization,''
{\it J.\ Artificial Intelligence Res.}, vol.~18, no.~1, pp.~45--81, 2003.

\bibitem{Temperley2009}
D.~Temperley,
``A Unified Probabilistic Model for Polyphonic Music Analysis,''
{\it J.\ New Music Res.}, vol.~38, no.~1, pp.~3--18, 2009.

\bibitem{Tanji2008}
M.~Tanji, D.~Ando and H.~Iba,
%M.~Tanji {\it et al.},
``Improving Metrical Grammar with Grammar Expansion,''
{\it Proc.\ AI 2008} (Springer LNAI 5360), pp.~180--191, 2008.

\bibitem{Tsuchiya2013}
M.~Tsuchiya, K.~Ochiai, H.~Kameoka and S.~Sagayama,
%M.~Tsuchiya {\it et al.},
``Probabilistic Model of Two-Dimensional Rhythm Tree Structure Representation for Automatic Transcription of Polyphonic MIDI Signals,''
{\it Proc.\ APSIPA}, pp.~1--6, 2013.

\bibitem{Takamune2014}
N.~Takamune, H.~Kameoka and S.~Sagayama,
%N.~Takamune {\it et al.},
``Automatic Transcription from MIDI Signals of Music Performance Using 2-Dimensional LR Parser (in Japanese),''
{\it Tech.\ Rep.\ SIGMUS}, vol.~2014-MUS-104, no.~7, pp.~1--6, 2014.

\bibitem{Nakamura2016}
E.~Nakamura, K.~Itoyama and K.~Yoshii,
``Rhythm Transcription of MIDI Performances Based on Hierarchical Bayesian Modelling of Repetition and Modification of Musical Note Patterns,''
{\it Proc.\ EUSIPCO}, pp.~1946--1950, 2016.

\bibitem{NakamuraSMC2016}
E.~Nakamura, K.~Yoshii and S.~Sagayama,
``Rhythm Transcription of Polyphonic MIDI Performances Based on a Merged-Output HMM for Multiple Voices,''
{\it Proc.\ SMC}, pp.~338--343, 2016.

\bibitem{Rabiner1989}
L.~Rabiner,
``A Tutorial on Hidden Markov Models and Selected Applications in Speech Recognition,''
{\it Proc.\ IEEE}, vol.~77, no.~2, pp.~257--286, 1989.

\bibitem{Heijink2000}
H.~Heijink, L.~Windsor and P.~Desain,
%H.~Heijink {\it et al.},
``Data Processing in Music Performance Research: Using Structural Information to Improve Score-Performance Matching,''
{\it Behavior Research Methods, Instruments, \& Computers}, vol.~32, no.~4, pp.~546--554, 2000.

\bibitem{Gingras2011}
B.~Gingras and S.~McAdams,
``Improved Score-Performance Matching Using Both Structural and Temporal Information from MIDI Recordings,''
{\it J.\ New Music Res.}, vol.~40, no.~1, pp.~43--57, 2011.

\bibitem{ICMC}
E.~Nakamura, Y.~Saito, N.~Ono and S.~Sagayama,
%E.~Nakamura {\it et al.},
``Merged-Output Hidden Markov Model for Score Following of MIDI Performance with Ornaments, Desynchronized Voices, Repeats and Skips,''
{\it Proc.\ Joint ICMC$|$SMC 2014}, pp.~1185--1192, 2014.

\bibitem{ISMIR2014}
E.~Nakamura, N.~Ono and S.~Sagayama,
%E.~Nakamura {\it et al.},
``Merged-Output HMM for Piano Fingering of Both Hands,''
{\it Proc.\ ISMIR}, pp.~531--536, 2014.

\bibitem{Kameoka2012}
H.~Kameoka, K.~Ochiai, M.~Nakano, M.~Tsuchiya and S.~Sagayama,
%H.~Kameoka {\it et al.},
``Context-Free 2D Tree Structure Model of Musical Notes for Bayesian Modeling of Polyphonic Spectrograms,''
{\it Proc.\ ISMIR}, pp.~307--312, 2012.

\bibitem{DemoPage}
E.~Nakamura, K.~Yoshii and S.~Sagayama.
{\it Demo Page of Polyphonic Rhythm Transcription}. (2016) [Online].
Available: \url{http://anonymous4721029.github.io/demo.html}, Accessed on: Dec.~20, 2016.

\bibitem{Conklin2002}
D.~Conklin,
``Representation and Discovery of Vertical Patterns in Music,''
in A.~Smaill et al.\ (eds.), {\it Music and Artificial Intelligence},
Lecture Notes in Artificial Intelligence, Springer, pp.~32--42, 2002.

\bibitem{Nakamura2015}
E.~Nakamura, N.~Ono, S.~Sagayama and K.~Watanabe,
%E.~Nakamura {\it et al.},
``A Stochastic Temporal Model of Polyphonic MIDI Performance with Ornaments,''
{\it J.\ New Music Res.}, vol.~44, no.~4, pp.~287--304, 2015.

\bibitem{Cont}
A.~Cont,
``A Coupled Duration-Focused Architecture for Realtime Music to Score Alignment,''
{\it IEEE Trans.\ on PAMI}, vol.~32, no.~6, pp.~974--987, 2010.

\bibitem{FactorialHMM}
Z.~Ghahramani and M.~Jordan,
``Factorial Hidden Markov Models,''
{\it Machine Learning}, vol.~29, pp.~245--273, 1997.

\bibitem{Raphael1999}
C.~Raphael,
``Automatic Segmentation of Acoustic Musical Signals Using Hidden Markov Models,''
{\it IEEE Trans.\ on PAMI}, vol.~21, no.~4, pp.~360--370, 1999.

\bibitem{Joder2011}
C.~Joder, S.~Essid and G.~Richard,
``A Conditional Random Field Framework for Robust and Scalable Audio-to-Score Matching,''
{\it IEEE Trans. on ASLP}, vol.~19, no.~8, pp.~2385--2397, 2011.

\bibitem{Krebs2015}
F.~Krebs, A.~Holzapfel, A.~T.~Cemgil and G.~Widmer,
``Inferring Metrical Structure in Music Using Particle Filters,''
{\it IEEE Trans. on ASLP}, vol.~23, no.~5, pp.~817--827, 2015.

\bibitem{Cemgil2000B}
A.~T.~Cemgil, B.~Kappen, P.~Desain and H.~Honing,
%A.~Cemgil {\it et al.},
``On Tempo Tracking: Tempogram Representation and Kalman Filtering,''
{\it J.\ New Music Res.}, vol.~29, no.~4, pp.~259--273, 2000.

\bibitem{Hashida2008}
M.~Hashida, T.~Matsui and H.~Katayose,
%M.~Hashida {\it et al.},
``A New Music Database Describing Deviation Information of Performance Expressions,''
{\it Proc.\ ISMIR}, pp.~489--494, 2008.

\bibitem{SalzerSchachter}
F.~Salzer and C.~Schachter,
{\it Counterpoint in Composition: The Study of Voice Leading}, Columbia University Press, 1989.

\end{thebibliography}
\end{document}